\documentclass{article}

\usepackage{arxiv}

\usepackage[utf8]{inputenc} 
\usepackage[T1]{fontenc}    
\usepackage{hyperref}       
\usepackage{url}            
\usepackage{booktabs}       
\usepackage{amsfonts}       
\usepackage{nicefrac}       
\usepackage{microtype}      
\usepackage{lipsum}         
\usepackage{graphicx}
\usepackage{natbib}
\usepackage{doi}
\usepackage{pythonhighlight}
\usepackage{longtable}

\usepackage{amsthm}
\theoremstyle{definition}
\newtheorem{definition}{Definition}[section]
\usepackage{epstopdf}
\usepackage{url}
\usepackage{xcolor}
\usepackage{wrapfig,graphicx,lipsum}
\usepackage{caption}
\usepackage[frozencache,cachedir=.]{minted}

\makeatletter
\AtBeginDocument{%
  \begingroup
  \normalsize
  \let\tmp@n@s\f@size
  \let\tmp@n@b\f@baselineskip
  \small
  \let\tmp@s@s\f@size
  \let\tmp@s@b\f@baselineskip
  \xdef\semismall@size{\fpeval{(\tmp@n@s+\tmp@s@s)/2}}%
  \xdef\semismall@baselineskip{\fpeval{(\tmp@n@b+\tmp@s@b)/2}}%
  \endgroup
}
\newcommand{\semismall}{\fontsize{\semismall@size}{\semismall@baselineskip}\selectfont}

\usepackage{subcaption}
\DeclareMathAlphabet{\msfsl}{OT1}{cmss}{m}{sl}

\usepackage{pythonhighlight}

\title{The S2 Hierarchical Discrete Global Grid as a Nexus for Data Representation, Integration, and Querying Across Geospatial Knowledge Graphs}

\newif\ifuniqueAffiliation
\uniqueAffiliationtrue

\ifuniqueAffiliation 
\author{ Shirly Stephen\\
	NCEAS | Department of Geography\\
	University of California\\
	Santa Barbara, CA, USA \\
	\texttt{shirlystephen@ucsb.edu} \\
	\And
	Mitchell Faulk \\
	NCEAS, University of California \\
	Santa Barbara, CA, USA \\
        \And
	Krzysztof Janowicz \\
	Geoinformatics, University of Vienna\\
	Vienna, Austria \\
        \And
        Colby Fisher \\
	Hydronos Labs \\
	New Jersey, USA \\
        \And
	Thomas Thelen \\
	NCEAS, University of California \\
        Santa Barbara, CA, USA \\
        \And
	Rui Zhu \\
	School of Geographical Sciences \\
        University of Bristol, Bristol, UK \\
	\And
	Pascal Hitzler \\
	Department of Computer Science \\
        Kansas State University, Kansas, USA \\
        \And
        Cogan Shimizu \\
        Department of Computer Science \& Engineering\\
	Wright State University, Ohio, USA \\
        \And
        Kitty Currier \\
        Department of Geography\\
	University of California, Santa Barbara\\
	California, USA \\
        \And
        Mark Schildhauer \\
        NCEAS\\
	University of California, Santa Barbara\\
	California, USA \\
        \And
        Dean Rehberger\\
        Department of History \\
        Michigan State University, MI, USA\\ 
        \And
        Zhangyu Wang\\
        Department of Geography\\
	University of California, Santa Barbara\\
	California, USA \\
        \And
        Antrea Christou \\
        Department of Computer Science \& Engineering\\
	Wright State University, Ohio, USA \\ Santa Barbara\\
	California, USA \\
}

\else
\usepackage{authblk}

\newbox{\orcid}\sbox{\orcid}{\includegraphics[scale=0.06]{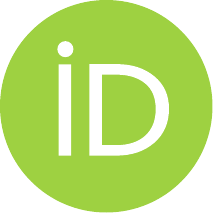}} 
\author[1]{%
	\href{https://orcid.org/0000-0000-0000-0000}{\usebox{\orcid}\hspace{1mm}David S.~Hippocampus\thanks{\texttt{hippo@cs.cranberry-lemon.edu}}}%
}
\author[1,2]{%
	\href{https://orcid.org/0000-0000-0000-0000}{\usebox{\orcid}\hspace{1mm}Elias D.~Striatum\thanks{\texttt{stariate@ee.mount-sheikh.edu}}}%
}
\affil[1]{Department of Computer Science, Cranberry-Lemon University, Pittsburgh, PA 15213}
\affil[2]{Department of Electrical Engineering, Mount-Sheikh University, Santa Narimana, Levand}
\fi

\renewcommand{\shorttitle}{S2 as a Nexus for Data Representation, Integration, and Querying Across GeoKGs}

\hypersetup{
pdftitle={The S2 Hierarchical Discrete Global Grid as a Nexus for Data Representation, Integration, and Querying Across Geospatial Knowledge Graphs},
pdfsubject={q-bio.NC, q-bio.QM},
pdfauthor={David S.~Hippocampus, Elias D.~Striatum},
pdfkeywords={First keyword, Second keyword, More},
}

\begin{document}
\maketitle

\begin{abstract}
Geospatial Knowledge Graphs (GeoKGs) have become integral to the growing field of Geospatial Artificial Intelligence. Initiatives like the U.S. National Science Foundation's Open Knowledge Network program aim to create an ecosystem of nation-scale, cross-disciplinary GeoKGs that provide AI-ready geospatial data aligned with FAIR principles. However, building this infrastructure presents key challenges, including 1) managing large volumes of data, 2) the computational complexity of discovering topological relations via SPARQL, and 3) conflating multi-scale raster and vector data. Discrete Global Grid Systems (DGGS) help tackle these issues by offering efficient data integration and representation strategies. The KnowWhereGraph utilizes Google's S2 Geometry --- a DGGS framework --- to enable efficient multi-source data processing, qualitative spatial querying, and cross-graph integration. This paper outlines the implementation of S2 within KnowWhereGraph, emphasizing its role in topologically enriching and semantically compressing data. Ultimately, this work demonstrates the potential of DGGS frameworks, particularly S2, for building scalable GeoKGs.
\end{abstract}

\keywords{geospatial knowledge graphs \and discrete global grid systems \and S2 geometry \and linked data \and query optimization \and spatial ontologies \and spatial data integration}

\section{Introduction}

Knowledge Graphs (KGs), originating from Semantic Web research \citep{hitzler2021swreview}, have revolutionized the data science landscape by facilitating the integration and enrichment of vast, multi-source, and heterogeneous data while empowering AI applications to deliver high-quality, trusted insights for scientific decision-making \citep{ji2021survey, abu2021domain}. To support translational research --- bridging the gap between fundamental research and real-world applications --- initiatives like the U.S. National Science Foundation's Open Knowledge Network (OKN) program\footnote{\url{https://www.proto-okn.net/}} and the Linked Open Data (LOD) Cloud\footnote{\url{https://lod-cloud.net/}} are driving the development of comprehensive ecosystems of nation-scale KGs that adhere to FAIR (Findable, Accessible, Interoperable, Reusable) data principles \citep{wilkinson2016fair}. These initiatives aim to create data-centric infrastructures for addressing complex cross-disciplinary challenges, ranging from climate change to economic equity. A central theme interlinking the various domain KGs in these infrastructures is the geospatial layer comprised of Geospatial Knowledge Graphs (GeoKGs), such as the KnowWhereGraph \citep{janowicz2022know} and Urban Flooding Open Knowledge Network (UF-OKN) \citep{johnson2022knowledge}. GeoKGs are now a critical component of modern Geospatial Artificial Intelligence (GeoAI), providing a structured and interconnected approach to managing and analyzing extensive spatial datasets from heterogeneous sources \citep{janowicz2022know, yan2020harnessing}. The Resource Description Framework (RDF) \citep{manola2004rdf} is the widely adopted paradigm for GeoKGs due to its semantic richness, high interoperability, and adherence to open standards. Additionally, the Open Geospatial Consortium's (OGC) GeoSPARQL standard \citep{car2022geosparql} enhances the representation of spatial information in RDF, enabling quantitative and qualitative spatial querying. This capability of GeoKGs is leveraged in AI systems to provide more intelligent, multi-faceted, and context-aware services \citep{janowicz2023fast}.

Despite the promise of GeoKGs, three primary challenges impede their effective implementation: handling and storing immense volumes of (geospatial) data, the computational complexity involved in large-scale spatial querying, and analyzing raster and vector data together at various spatial scales. Graph databases capable of managing large datasets (e.g., Neptune, Neo4j) lack any GeoSPARQL support. GeoSPARQL-compliant RDF databases \citep{jovanovik2021geosparql} like Blazegraph\footnote{\url{https://blazegraph.com/}}, GraphDB\footnote{\url{https://graphdb.ontotext.com/}}, Virtuoso\footnote{\url{https://virtuoso.openlinksw.com/}}, and Stardog\footnote{\url{https://www.stardog.com/}} use indexing methods such as R-trees, quadtrees, and geohashing. However, these methods are not optimized for cross-scale analyses, global datasets, and dense graphs with high degrees of spatial overlap \citep{huang2019assessment, theocharidis2019srx}. High memory costs from building and maintaining extensive indexes, computational costs of traversing deep R-trees and quadtrees \citep{kothuri2002quadtree}, and inefficiencies in proximity searches due to boundary ambiguities in geohashing \citep{geohash} are just some challenges with large graphs. While these indexing methods can accelerate fundamental spatial filtering queries such as point-based queries, containment queries, and spatial joins for range and nearest-neighbor queries, complex queries involving spatial refinements, such as polygon intersection, spatial cross-matching, and identifying spatial patterns, remain highly expensive. For example, queries like ``\emph{Find the regions where earthquakes occurred in 2023}'' or ``\emph{Find comparable areas that are earthquake-prone}'' require a large number of spatial joins to determine relevancy, and can be inefficient or even return incorrect results \citep{huang2019assessment}. Experimental evaluations show that topological computation becomes highly expensive or nearly intractable in GeoKGs characterized by extensive data volume, uneven data scale, and diverse categories of data \citep{theocharidis2019srx, li2022performance}. Furthermore, the qualitative spatial functions that GeoSPARQL provides are insufficient for multi-scale analytics \citep{manley2021scale}. For instance, conflating a vector region dataset with a map scale of 1:50,000 alongside a land-use raster dataset of 30 m resolution can be problematic \citep{davis1991environmental}. Addressing these challenges requires innovative data integration and management strategies that can efficiently handle large, multi-scale, multi-format geospatial data and support complex topological queries in the context of GeoKGs.

Discrete Global Grid Systems (DGGS) are spatial indexing frameworks that tessellate the Earth's surface into hierarchical, regular grids \citep{sahr2003geodesic}. Unlike traditional GIS systems that use latitude and longitude coordinates, a DGGS employs cell identifiers (cell IDs) to uniquely reference discrete portions of the Earth's surface. DGGS frameworks have been utilized in geospatial applications since the early 1990s \citep{goodchild1994geographical}, but have more recently emerged as a key technology for next-generation GIS \citep{li2020geospatial, hojati2022giscience, bondaruk2020assessing} and Digital Earth systems \citep{mahdavi2015survey, robertson2020integrated, yao2019enabling, hojati2022giscience}. This is primarily driven by the standardization of DGGS configurations by the OGC in 2017 \citep{ogcdggs} that spurred the proliferation of DGGS implementations and their open-source software libraries, such as Google's S2 Geometry \citep{veach2017s2}, Uber's H3 \citep{h3}, rHEalPIX \citep{gibb2016rhealpix}, and OpenEAGGR \citep{eaggr}. These libraries provide (quantization) functions to assign geometries to cells, and spatial operations for cell navigation \citep{li2020geospatial}. Their algorithms leverage cell IDs as well as their inherent inter-relations to replace complex spatial computations with simple coding operations, making DGGS frameworks powerful tools for efficient data integration, retrieval, and reduction of analytical complexity across various geospatial application domains \citep{li2021integration, kranstauber2015global, romanov2018emissivity, pereira2024ride, lin2018discrete}. A seminal paper by Goodchild in 2018 \citep{goodchild18} highlighted the benefits of DGGS for multi-scale, multi-format data integration within linked data frameworks, sparking interest in the GeoKG and OKN communities \citep{bastrakova2020enabling}. Despite its potential, the adoption of DGGS in GeoKGs remains limited. One reason is the lack of support in graph databases for DGGS library interfaces or full support for gridded representations of geospatial data. While GeoSPARQL 1.1 \citep{geosparql} introduces a generic format for DGGS geometry serialization \citep{car2022geosparql}, it does not yet provide the ability to interpret specific cell IDs according to the various DGGS frameworks available. Furthermore, testing this implementation in RDF databases is still in its early stages \citep{habgood2022implementation}. Few graph databases support DGGS-based indexing, such as those using S2 and H3 (e.g., NebulaGraphDB), but these are not RDF-based or GeoSPARQL-compliant \citep{jovanovik2021geosparql}. The commercial Foursquare GeoKG \citep{fsquare} utilizes H3 for location-specific business use cases, but only as an indexing framework. More sophisticated usages of DGGS in GeoKGs, such as for spatial data compression through semantic compression have been postulated \citep{zalewski2021semantic} but not practically explored until KnowWhereGraph.

The KnowWhereGraph (KWG) \citep{janowicz2022know}, a GeoKG within the OKN framework, employs the S2 Geometry to spatially integrate large-scale geospatial data by optimizing their storage, processing, visualization, and analysis across various data types, shapes, scales, and precision. KWG transforms siloed heterogeneous location and environmental observation data into actionable insights for environmental intelligence by \emph{enhancing cross-domain data interoperability}, \emph{contextualizing data}, and \emph{generating AI-ready geospatial data}. Rather than relying on S2 as an implicit indexing method in GraphDB (the graph database where KWG is deployed), KWG represents S2 cells as a set of nodes within an S2-RDF graph. GeoSPARQL's Dimensionally Extended 9-Intersection Model (DE-9IM) relations \citep{egenhofer19949} are employed to model cell navigation (parent, child, neighbor) and topological connections between S2 cells and other geographic features. Integrating S2 as a spatial reference framework within KWG provides a robust infrastructure that supports cross-domain data linkages and abstracts spatial analytics while fostering a rich environment for modeling and computation of geographical information.

This paper details this implementation by outlining the development of the S2-RDF Graph and describing the following two methods to quantize spatial data onto S2 cells.
\vspace{-7pt}
\begin{enumerate}
    \item \emph{Topological enrichment of cells} involves materializing topological relations (based on DE-9IM) between S2 cells and geographic features having explicit geometries in the graph. This approach mitigates the tractability challenges of calculating these relationships at runtime via GeoSPARQL, even when utilizing efficient implicit spatial indexing methods.
    
    \item \emph{Grid-based data discretization} involves transforming non-DGGS data (both vector and raster) using techniques like statistical aggregation, decomposition, and spatial overlay to represent them as S2-cell-based gridded data. This process enables geometric simplification, scalable data representation, and integrated raster/vector analysis. In this framework, the S2 cells serve as the spatial $\emph{feature of interest}$ \citep{janowicz2019sosa}, linking various thematic observations.
\end{enumerate}
\vspace{-7pt}

The significant spatial computational capabilities of the S2 Geometry library \citep{veach2017s2} are utilized in a Python-based RDF processing environment to materialize statements for enrichment and discretization. Through topological enrichment, S2 cells are tagged with spatial features, thereby associating cells with observations linked to those features, e.g., S2 cells are linked to climate observations via topologically connected climate division features. Through discretization, S2 cells are enriched with temporally-scoped observations computed for the specific cell area, e.g., the area of a crop type within a cell. Once the S2 grid is augmented with each dataset using these methods, each cell node can be queried as a set of individual objects containing diverse pieces of information that can then be \emph{spatially fused} or \emph{conflated} for analysis. This approach is powerful because once a dataset is quantized on the S2 grid, it eliminates the need for repeated spatial analytics and related GIS processes to integrate and compare disparate datasets. Our experience indicates that leveraging S2 in KWG has enabled efficient multi-source data processing and faster qualitative spatial querying while constructing an enriched GeoKG that augments graph embeddings \citep{iliakis2022geospatial} and cross-disciplinary data integration within the OKN framework (Section~\ref{cross-integration}).

This paper aims to provide guidelines and examples for implementing and utilizing S2 in GeoKGs, highlighting its analytical advantages. By doing so, it seeks to advance GeoAI and offer a scalable solution for managing and analyzing large-scale geospatial data within GeoKGs.

The rest of the paper is structured as follows: Section~\ref{sec:background} provides background on GeoKGs and DGGS frameworks, highlighting KWG and S2 Geometry. Section~\ref{sec:methodology} introduces the workflow for modeling and ingesting S2 data into KWG, along with the technical strategy for quantization. Section~\ref{sec:implementation} discusses the implementation of the two quantization methods in KWG. Section~\ref{sec:usage} evaluates the performance and benefits of S2 integration through various use cases, demonstrating improvements in data processing, query efficiency, and scalability. Finally, Section~\ref{sec:conclusion} discusses the broader implications of using DGGS, particularly S2 Geometry, within GeoKGs, including achieved benefits, current limitations, and areas for future research.

\section{Theoretical Background and Related Research}\label{sec:background}

\subsection{Geospatial Knowledge Graphs}
Geospatial knowledge graphs (GeoKGs) emerged in the late 2000s as open data ecosystems, primarily aiming to democratize data access in alignment with FAIR (Findable, Accessible, Interoperable, Reusable) data principles \citep{wilkinson2016fair}. Early graphs such as GeoNames\footnote{\url{https://www.geonames.org/}} and LinkedGeoData\footnote{\url{http://linkedgeodata.org/}} predominantly comprised \emph{explicit geographic} entities, such as named places, natural features, landmarks, roads, and historical features, with point geometries and limited semantics. In recent years, GeoKGs have expanded in scope to model \emph{geographically themed} data across many domains, such as from environmental \citep{janowicz2022know, zhu2021environmental}, public health \citep{jiang2020interactive, zhu2022covid}, disaster management \citep{janowicz2022know}, transportation \citep{bockling2024planet, qi2023evkg}, production logistics \citep{zhao2022digital}, and gaming domains \citep{jiang2020interactive} demonstrating the versatility and impact of GeoKGs in various application areas. These GeoKGs play a pivotal role in cross-domain geospatial analytics \citep{janowicz2022know}, geospatial question answering \citep{mai2021geographic, weinberger2022towards}, and geo-visualization \citep{balla2020geovisualization, li2023geographvis}. Moreover, their rich contextual information makes them essential training data for spatial models, facilitating the development of spatial and temporal reasoning \citep{zhu2022reasoning}, predictive analytics, and decision support systems \citep{zhu2022reasoning, mai2022symbolic}. Examples of such rich GeoKGs in the OKN ecosystem \citep{okn} include the 
KWG\footnote{\url{https://knowwheregraph.org/}} that focuses on environmental intelligence \citep{janowicz2022know}; the UF-OKN\footnote{\url{https://ufokn.com/}} for flood prediction, response, mitigation, and prevention \citep{saksena2022urban}; and the Safe Agricultural Products and Water Graph\footnote{\url{https://sawgraph.github.io/}} (SAWGraph) to understand per- and polyfluoroalkyl substances (PFAS) contamination in food and water systems.

\vspace{7pt}
\noindent \textbf{\emph{A. Preliminaries of the RDF Data Model: }} The structure and principles of RDF and the Web Ontology Language (OWL) \citep{hitzler2009owl} directly influence how S2 is used to synthesize, integrate, and query data within KWG, as will be discussed later in Section~\ref{sec:implementation}. To facilitate understanding, this section briefly introduces the basic components and structure of RDF and SPARQL. For a more comprehensive introduction to RDF and SPARQL, refer to \citep{manola2004rdf,hitzler2010fost,world2013sparql}.

\begin{definition}
\emph{RDF triple}: An RDF triple $\mathit{t(s,p,o)}$ is the basic atomic entity of a knowledge graph. It expresses a single statement about semantic data. The subject (s) and object (o) are considered graph vertices referring to a resource or a simple value (called literal), and the property (p) (a.k.a. predicate) is the directed edge that connects the two vertices. The \emph{resources} in $\mathit{t(s,p,o)}$ are given uniform resource identifiers (URI) or are blank nodes (denoting an unknown resource).
\end{definition}

\begin{definition}
\emph{RDF graph}: An RDF graph is a collection of RDF triples. Such a graph can essentially be viewed as a directed graph $\mathit{G = (V, E)}$ where $\mathit{V}$ denotes the set of vertices that can be resources or literals, and $\mathit{E}$ denotes the set of directed edges. 
\end{definition}

\begin{figure}[h]
    \centering
    \includegraphics[width=0.6\linewidth]{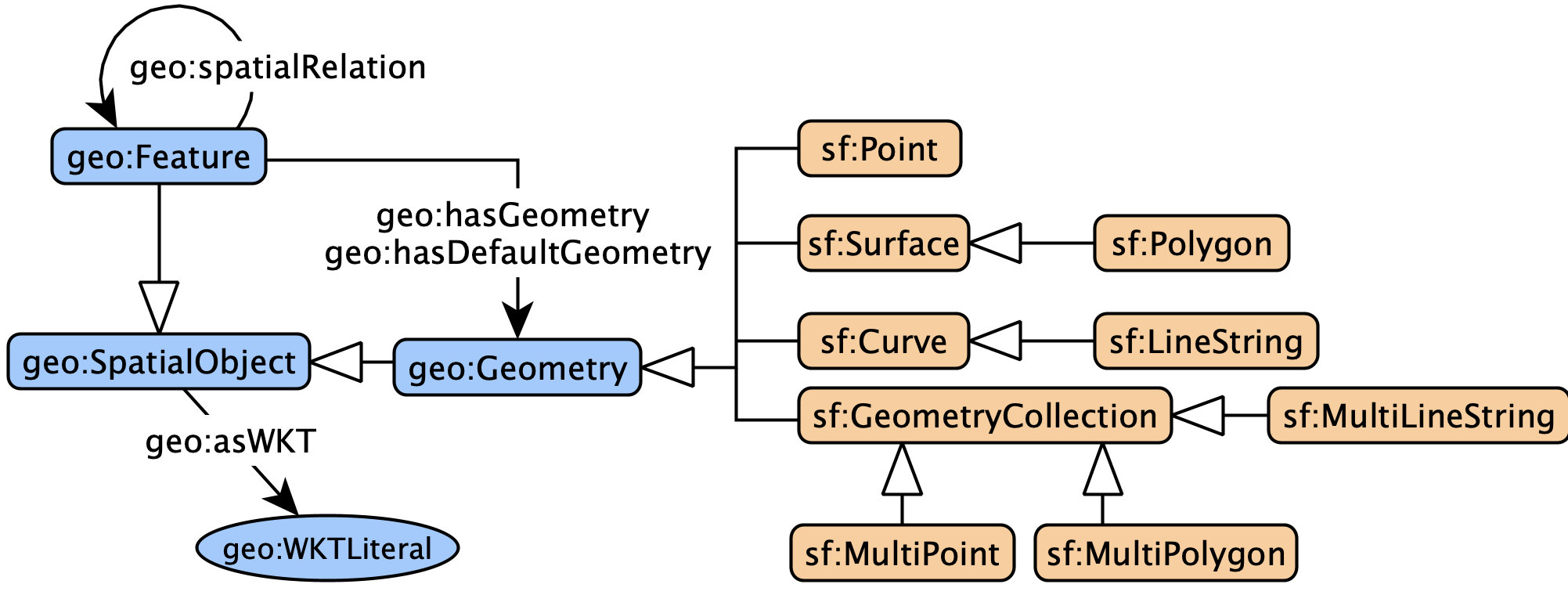}
    \caption{GeoSPARQL's core concepts are shown in blue boxes, and Simple Features geometry classes in beige boxes. White-headed arrows represent $\texttt{rdfs}$:$\texttt{subClass}$ relations, while filled arrows denote object or data properties}
    \label{fig:geosparql}
\end{figure}

\begin{definition}
\emph{SPARQL query}: A SPARQL query $\mathit{Q}$ is defined as tuple $\mathit{Q =(E,G,R)}$. $\mathit{E}$ denotes the algebraic expression built from graph patterns and solution modifiers. $\mathit{G}$ is the RDF graph being queried. $\mathit{R}$ denotes the result form, such as SELECT, CONSTRUCT, DESCRIBE, or ASK, which specifies how the query results are processed and presented. The algebraic expression $\mathit{E}$ can include various graph patterns and solution modifiers like PROJECTION, DISTINCT, LIMIT, or ORDER BY. The simplest graph pattern in SPARQL is the triple pattern, which matches triples in the RDF graph.
\end{definition}
\noindent \textbf{\emph{B. GeoSPARQL:}}
In the RDF domain, OGC's GeoSPARQL standard \citep{car2022geosparql} is the leading specification for representing and querying geospatial data. It is implemented as an extension by many RDF graph databases \citep{jovanovik2021geosparql}. GeoSPARQL leverages OGC's Simple Features ontology\footnote{\url{https://www.ogc.org/standard/sfa/}} for defining spatial entities as shown in Figure.~\ref{fig:geosparql}. The $\texttt{geo:SpatialObject}$ class includes any resource that can have a spatial representation. The $\texttt{geo:Feature}$ class specifically represents spatial objects with concrete geographical shapes, linked to their geometries (serialized as GML or WKT) via the $\texttt{geo:hasGeometry}$ property. GeoSPARQL supports qualitative relations such as containment and overlap for evaluating topological relationships between entities, with $\texttt{geo:spatialRelation}$ subsuming specific properties. Additionally, the specification provides properties for non-topological spatial operations (e.g., distance, buffer) and spatial aggregation functions. GeoSPARQL can be used in conjunction with OWL to develop detailed ontologies that enhance and constrain spatial data.
\vspace{7pt}

\noindent \textbf{\emph{C. Spatial Query Challenges in GeoKGs:}}
The size of GeoKGs can increase significantly when handling complex geometries, such as multi-polygons and polylines. For instance, the geometric representation of Greece from GADM\footnote{\url{https://gadm.org/data.html}} is a multi-polygon consisting of 695 individual polygons, with approximately 29,900 vertices. This results in a Well-Known Text (WKT) representation that is about 9MB in size. High precision can adversely affect the graph's performance in terms of storage, indexing, and query efficiency \citep{regalia2017revisiting}, while reducing precision can impact the quality of spatial querying. Algorithms determining topological relations using geometries operate with polynomial time complexity relative to the number of nodes in the geometries being tested \citep{rigaux2002spatial}. Graph databases that implement popular space-dividing indexing mechanisms, such as quadtrees and geohashing (e.g., GraphDB), still encounter limitations in spatially pruning search spaces. Geohashing can split regions of interest across multiple cells, leading to boundary issues where a single spatial entity is divided by cell boundaries. This affects the accuracy and efficiency of spatial queries, as additional computation is needed to merge or handle entities spanning multiple geohashes \citep{papadias2003query}. This is particularly problematic for linear features (e.g., rivers) or extensive areas (e.g., country boundaries). Each geohash string is independent, lacking a hierarchical relationship between different levels of resolution, and they are not ordered based on spatial proximity. This non-intuitive ordering complicates range queries and neighborhood searches, especially for features such as political borders that often span multiple geohash cells \citep{geohash}. Quadtrees provide adaptive resolution indexing capabilities and improve query performance by reducing the number of nodes that need to be inspected. However, high-resolution data can result in very deep quadtree structures, which can lead to performance bottlenecks, particularly in large-scale graphs \citep{quad}. Experimental evaluations have also shown that spatial joins for complex geometries are not optimized in GeoSPARQL-compliant RDF databases \citep{huang2019assessment, ioannidis2021evaluating, li2022performance}. These size-related query challenges of GeoKGs re-emphasize prior statements \citep{regalia2019computing} that storing accurate representations of geospatial data in their original vector or raster format, including complex geometries as RDF literals, while reasonable, may not be suitable for storage and query efficiency.

\subsection{The KnowWhereGraph}\label{sec:kwg}
Envisioned as a \emph{gazetteer of gazetteers} that links heterogeneous data via joint place identifiers, KnowWhereGraph (KWG) empowers decision-makers with on-demand and comprehensive location insights derived from a diverse range of data. By minimizing data processing overhead, KWG provides semantically rich, contextualized, and analysis-ready data for environmental intelligence analytics, with a particular focus on two key use cases. The humanitarian relief use case \citep{zhu2021providing} is tailored to aid humanitarian organizations in swiftly identifying and mobilizing relevant personnel with suitable expertise to respond effectively to imminent or ongoing disasters. The farm-to-table use case \citep{janowicz2022know} aims to address concerns regarding the safety of crops affected by smoke and ash from wildfires, providing crucial insights for stakeholders along the supply chain. The graph is implemented in GraphDB, an RDF graph database that supports SPARQL 1.1 and GeoSPARQL. The SPARQL endpoint to the graph is \url{https://stko-kwg.geog.ucsb.edu/graphdb/sparql}.

\begin{figure}[!ht]
    \centering
    \includegraphics[width=1.0\linewidth]{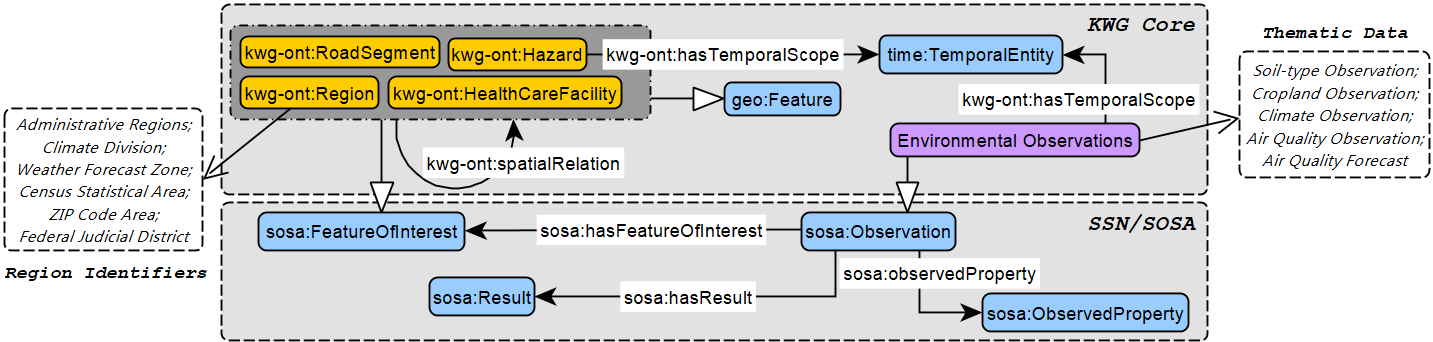}
    \caption{Schema diagram that denotes the core concepts/themes in KWG (explicit geographic features denoted using yellow boxes and geographically themed observations denoted using the purple box) and their extension of SSN/SOSA, GeoSPARQL, and OWL-Time ontologies (denotes using blue boxes). The SSN/SOSA pattern denotes the generic graph to query KWG observations. White-headed arrows denote $\texttt{rdfs}$:$\texttt{subClass}$ relationships.}
    \label{fig:kwg-core}
\end{figure}

KWG includes data about a wide range of \emph{region identifiers}, encompassing global administrative regions, climate divisions, weather forecast zones, census statistical areas, federal judicial districts, and zip code areas. Other feature data include hazard events, road segments, and public health departments. The graph also integrates a set of geographically themed data that significantly influences the use cases. These encompass observation and measurement data, such as climate observations, air quality indices, public health metrics, and natural hazard impacts. Figure~\ref{fig:kwg-core} is an abstracted pattern of the KWG Ontology \citep{kwgontology2023} depicting how: 1) entities with explicit geometries are modeled using GeoSPARQL, and 2) environmental observations are modeled using the SSN/SOSA ontology \citep{janowicz2019sosa,haller2019modular}. At the heart of this pattern is the core class \texttt{kwg-ont:Region}, which encapsulates dataset-specific subclasses of regions. Qualitative spatial relations or simply topological relations between features are established using the set of DE-9IM relations denoted as $\mathcal{T} =$ $\{ \texttt{kwg-ont:sfWithin}, \texttt{kwg-ont:sfContains}, \texttt{kwg-ont:sfTouches}, \texttt{kwg-ont:sfCrosses}, \texttt{kwg-ont:sfOverlaps}, \\ \texttt{kwg-ont:sfIntersects} \}$. Relations in $\mathcal{T}$ are sub-properties of the generic $\texttt{kwg-ont:spatialRelation}$ within the \texttt{kwg-ont} namespace. Each relation follows the naming convention of its topological counterparts in GeoSPARQL, e.g., $\texttt{kwg}$-$\texttt{ont}$:$\texttt{sfContains}$ is semantically analogous to $\texttt{geo}$:$\texttt{sfContains}$. However, the spatial relations in the \texttt{kwg-ont} namespace are decoupled from GeoSPARQL's relations (i.e., they are not axiomatically related) to prevent interference with inferencing in the latter's inbuilt functions. Much of the observation data in KWG are not feature-based but have spatial information encoded through identifiers such as ZIP code, FIPS code, and geoid. These identifiers allow thematic data to be linked to their associated region identifiers. Observation data are georeferenced to their corresponding geographical feature of interest using SOSA's $\texttt{sosa:hasFeatureOfInterest}$ property as shown in Figure~\ref{fig:kwg-core}. This observation-centric linking paradigm is uniformly applied across all data with a spatial dimension \citep{janowicz2012observation}. 

\vspace{10pt}

Regions are spatially cross-integrated, allowing thematic data to also be cross-linked. For instance, census statistics or climate observations can be analyzed together, but this can be conflicting since different types of regions are of different scales but also discretized for distinct purposes (e.g., weather forecast zones are delineated based on differences in weather, while census statistical areas are delineated based on the number of inhabitants and urbanization). By using the S2 Geometry all the disconnected themes and layers are harmonized onto a common discretized spatial unit of analysis that is unbiased by natural and human processes. This quantization of KWG data onto S2 cells will be discussed later in Section~\ref{sec:implementation}.

Table~\ref{tab:statistics} provides an overview of KWG graph statistics (in terms of the number of total triples, feature nodes, geometry nodes, and observation nodes). The number of complex geometries (polylines/multi-polygons) is an indication of the size and complexity of the graph.

\begin{table}[]
\centering
\begin{tabular}{|l|l|}
\midrule
Total number of triples in the graph &  28,643,164,592 \\ \midrule
Number of regions (i.e., instances of $\texttt{kwg}$-$\texttt{ont}$:$\texttt{Region}$) in the ABox &  438,927 \\ \midrule
Number of features (i.e., instances of $\texttt{geo}$:$\texttt{Feature}$) in the ABox &  48,397,918 \\ \midrule
Number of features of interest ($\texttt{sosa}$:$\texttt{FeatureOfInterest}$) in the ABox &  6,705,844 \\ \midrule
Number of geometries (i.e., instances of $\texttt{geo}$:$\texttt{Geometry}$) in the ABox &  48,077,280 \\ \midrule
Number of point geometries (i.e., instances of $\texttt{sf}$:$\texttt{Point}$) in the ABox &  8,424,208 \\ \midrule
Number of lines (i.e., instances of $\texttt{sf}$:$\texttt{LineString}$) in the ABox &  741,214  \\ \midrule
Number of simple polygon geometries (i.e., instances of $\texttt{sf}$:$\texttt{Polygon}$) in the ABox &  38,546 \\ \midrule
Number of complex polygon geometries (i.e., instances of $\texttt{sf}$:$\texttt{MultiPolygon}$) in the ABox &  37056438 \\ \midrule
Number of observations (i.e., instances of $\texttt{sosa}$:$\texttt{Observation}$) in the ABox &  254,006,244 \\ \midrule
\end{tabular}
\caption{\normalsize KnowWhereGraph (v 3.0 or KWG - Santa Barbara) statistics (from 09/15/2023).}
\label{tab:statistics}
\end{table}

\subsection{Discrete Global Grid Systems}\label{sec:dggs}

\begin{figure}[h]
    \centering
    \begin{subfigure}[t]{0.14\textwidth}
        \includegraphics[width=\textwidth]{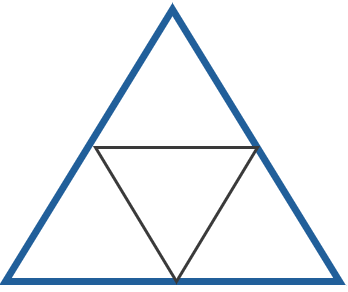}
        \caption{{\semismall e.g., OpenEAGGR}}
        \label{fig:triangle}
    \end{subfigure}
    ~\qquad \qquad \qquad
    \begin{subfigure}[t]{0.12\textwidth}
        \includegraphics[width=\textwidth]{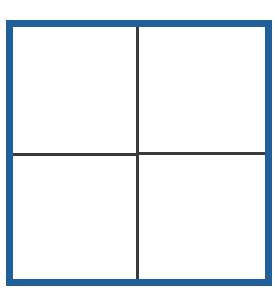}
        \caption{{\semismall e.g., S2}}
        \label{fig:square}
    \end{subfigure}
    ~\qquad \qquad \qquad
    \begin{subfigure}[t]{0.12\textwidth}
        \includegraphics[width=\textwidth]{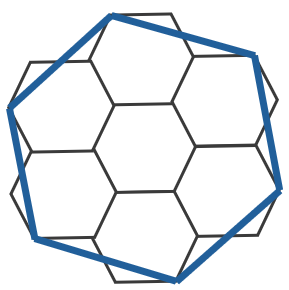}
        \caption{{\semismall e.g., H3}}
        \label{fig:hexagon}
    \end{subfigure}
    \caption{Congruence (i.e., parent-child containment relationships), and aggregation or decomposition resolutions (e.g., squares can be aggregated in groups of four to form coarser resolution objects) in DGGS with varying polyhedron shapes.}\label{fig:dggs}
\end{figure}

A Discrete Global Grid (DGG) is a spatial reference system that tessellates the Earth's surface into a series of discrete, connected, and well-aligned cells. A DGGS is a hierarchical composition of multi-resolution DGGs. At the core of a DGGS are the concepts of cells, levels, and coverings. A \emph{cell} is a fundamental unit in a DGG and represents a specific geographic region with a fixed location and area. Each DGG layer corresponds to a \emph{level} representing the granularity of cells in that layer. Cell sizes and cell counts vary between levels. Each cell at a parent level is linked to a cell at the next level with a finer predefined resolution, allowing for efficient hierarchical operations. Cells are uniquely identified by stable cell identifiers (cell IDs) that provide information regarding the hierarchical level, and structural connections to parents and neighboring cells. The collection of cells that covers a given region or shape is a \emph{covering}. A DGGS is \emph{congruent} if and only if each cell at resolution $\mathit{k}$ consists of a union of cells at resolution $\mathit{k+1}$. A DGGS is \emph{aligned} if and only if each point within a cell at resolution $\mathit{k}$ is also with its child cell at resolution $\mathit{k+1}$ \citep{sahr2003geodesic}. The congruence and alignment properties of cells differ between various DGGSs (see a comparison presented in Figure~\ref{fig:dggs}).

\vspace{8pt}

The design of a DGGS is primarily governed by the base polyhedron (cube, dodecahedron, icosahedron, octahedron, and tetrahedron) adopted for the spherical approximation of the Earth, and the method adopted to partition the polyhedron into finer cells. The latter choice determines the cell type (square, triangle, or hexagon), aggregation or decomposition resolutions, and congruence properties as compared in  Figure~\ref{fig:dggs}. Other design choices such as methods of orienting the polyhedron relative to the surface of the Earth, and methods to transform the partitioned planar cells to the sphere are equally important as they determine the level of area, shape, and angular distortion that can impact downstream spatial analysis. DGGSs primarily use hierarchy-based, coordinate-based, or space-filling curve-based indexing systems \citep{mahdavi2019geospatial}. DGGS frameworks that use a space-filling curve, for instance Hilbert curve (adopted in S2), and Gosper curve (adopted in H3), provide an explicit structural ordering of cells along the path of the curve. A cell's position (or coordinates) along this curve uniquely determines its cell ID, and resolution. Space-filling curves map $n$-dimensional spatial data to a one-dimensional sequence of cell IDs, approximately preserving the spatial locality of the data points \citep{uher2019hierarchical}. This transformation provides an efficient method for indexing and querying geographic information. Different DGGSs offer different performance capabilities for different objectives, e.g., rHEALPix for latitudinal data analysis \citep{bowater2019isolatitude} and area-based statistics \citep{gibb2016rhealpix}, H3 for modeling of movements through the grid \citep{kmoch2022area}, and S2 for smoother multi-resolution analysis \citep{kmoch2022area}.

The OGC's DGGS standard \citep{ogcdggs} specifies three core operational requirements for DGGS specifications: quantization, spatial relation, and interoperability operations. \emph{Quantization operations} assign raw or synthesized vector and raster spatial data to DGGS cells. \emph{Spatial relation operations} utilize the hierarchical and connected structure of cells and cell typologies to facilitate cell traversal and spatial analysis functions. Interoperability operations are designed to communicate with end-users or other spatial data infrastructures via standard APIs and data formats. Most DGGSs have open-source software libraries as well as bindings in various programming languages that provide methods for these operations. For example, the H3 library\footnote{\url{https://h3geo.org/}} is implemented in C and the S2 library\footnote{\url{https://s2geometry.io/}} in C++. Existing works have reviewed the operations supported by the state-of-the-art DGGS libraries, their adherence to the OGC standard \citep{li2020geospatial}, and trade-offs of their different characteristics designs \citep{bondaruk2020assessing, kmoch2022area}. 

\vspace{4pt}
\textbf{\emph{DGGS in Traditional Geospatial Applications:}} In the last two decades, DGGSs have been instrumental in actualizing the Digital Earth vision \citep{goodchild2000discrete} through their implementations in sensor data integration \citep{birk2018design}; multi-dimensional vector big data management \citep{robertson2020integrated, sirdeshmukh2019utilizing}; storing pre-aggregated results \citep{birk2018design, robertson2020integrated}; advanced multi-resolution geo-visualization \citep{raposo2019geovisualization, li2022geovisualization}; incorporating spatial uncertainty in big data analysis \citep{robertson2020integrated}; and storing, fusing, and rendering raster data \citep{strassburg2010global, li2021integration,rawson2022intelligent, bousquin2021discrete, miao2023big}. Specific application domains implementing DGGS include terrain data modeling \citep{li2021integration, lin2018discrete}, ecological studies \citep{kranstauber2015global, birch2007rectangular}, environmental observations modeling \citep{lin2018discrete, romanov2018emissivity}, ride-hailing services \citep{pereira2024ride}, and geo-social networking \citep{wozniak2021hex2vec}. 

\vspace{4pt}
\textbf{\emph{DGGS in GeoKGs:}} More recently DGGSs have been adopted in GeoKGs for many disciplines and applications. The ISEA3H DGGS is adopted to provide a standardized representation of national multi-source terrain data in \citep{li2021integration}. Elevation data from multiple raster datasets are quantized on the grid by resampling with bilinear interpolation at the cell centroid locations. \citep{han2022emergency} uses the GeoSOT-3D DGGS to integrate spatiotemporal information in the airspace environment for selecting airport sites in response to emergencies. The Spatial-Temporal Knowledge Graph (STKG) in \citep{bockling2024planet} adopts KWG's methodology to transform Open Street Map (OSM) data onto the H3 grid by topologically linking OSM geometries to grid cells via spatial predicates defined by GeoSPARQL. However, the relations between the grid cells use a set of non-topological relations. Due to the non-congruency of H3 cells, inferring spatial relations across multiple resolutions is not straightforward. This necessitates precomputing the links between OSM geometries and grid cells at several levels, resulting in a very large graph. AugGKG is another example that integrates spatio-temporal data on the GeoSOT for spatio-temporal question answering by creating sub-knowledge graphs on each time slice \citep{han2023auggkg}. W{\=a}hi, a discrete global grid gazetteer \citep{adams2017wahi} links places in the GeoNames database to each level of three DGGS (ISEA3H, ISEA4H, ISEA4T). Links were created using two types of spatial footprints of places in the gazetteer. The first type is defined by the grid cells that spatially intersect with the source geometry, which can be polygons (when available) or points, and will always cover the source polygon. The second type is defined by the spatial intersection of the centroids of grid cells with geographic features represented by polygons. If no centroid intersects, the single cell with the closest centroid is matched to the feature. This second type of footprint is always smaller or equal in size to the first type and is not guaranteed to cover the entire source polygon. Due to the current unavailability of the server hosting W{\=a}hi, we were unable to determine the exact predicate used to bin point data into cells. The data is stored in a PostgreSQL PostGIS database, with the option to export data into GeoJSON-LD format.

\subsection{The S2 Discrete Global Grid}\label{sec:s2}
\begin{figure}[h]
    \centering
    \begin{subfigure}[b]{0.3\textwidth}
        \includegraphics[width=\textwidth]{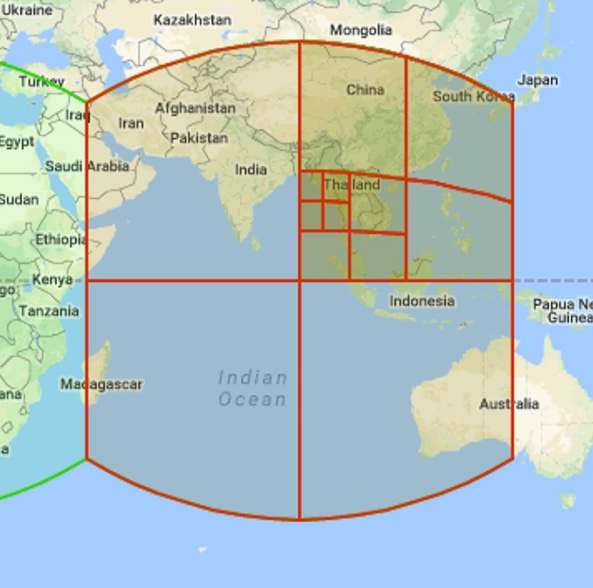}
        \caption{}
        \label{fig:s2dggs}
    \end{subfigure}
    ~\qquad \qquad
    \begin{subfigure}[b]{0.37\textwidth}
        \includegraphics[width=\textwidth]{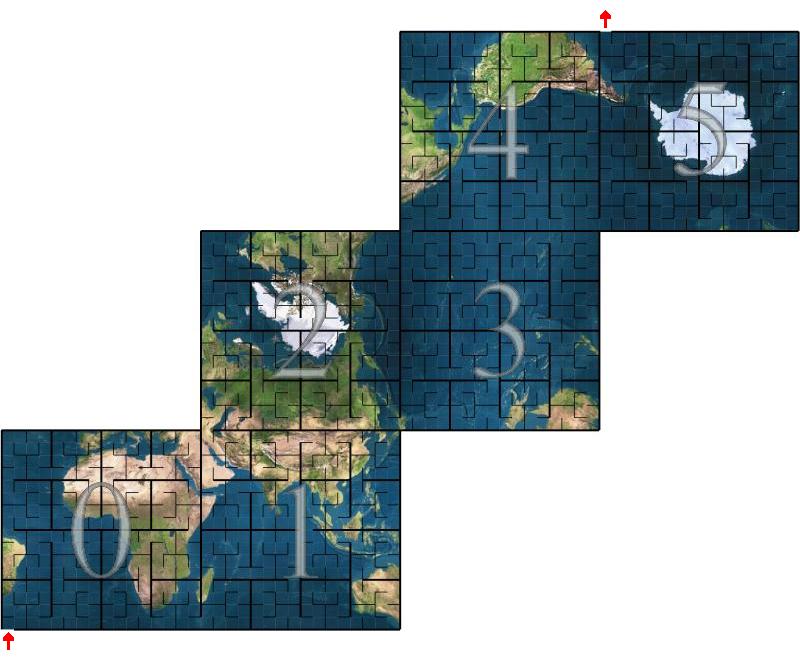}
        \caption{}
        \label{fig:regionCoverer}
    \end{subfigure}
    \caption{\textbf{(a)} Hierarchical tessellation of the Earth using the S2 Geometry. \textbf{(b)} The S2Cell hierarchy projecting the Earth onto six ``base cells''. These illustrations are adopted from the S2 Library documentation  \protect\citep{veach2017s2}.}\label{fig:s2}
\end{figure}

Google's S2 Geometry defines a sophisticated system for tessellating the Earth into a hierarchical mosaic of S2 cells, which are specialized spherical quadrilaterals (Figure~\ref{fig:s2dggs}). The edges of these cells are geodesics, meaning they follow the shortest path along the sphere's surface. This hierarchical mosaic is constructed through a nested structure, where each cell acts as a node within a quadtree, and each non-child node can be recursively subdivided to form four child cells. S2 cells are sequentially indexed on a Hilbert space-filling curve that traverses the unit sphere's surface projected onto the six faces of the cube using the Quadrilateralized Spherical Cube (QSC) projection method. This projection ensures a relatively even distribution of cells across the sphere's surface, maintaining geometric consistency and balance. Each S2 cell is uniquely identified by a 64-bit \pyth{S2CellID}, which encodes the cell's location on the curve and its level within the S2 hierarchy. The average area of cells within each level ranges from approximately $8.5$x$10^7 km^2$ at level 0 to about $0.74 cm^2$ at level 30. 


For quantization operations and semantic compression in GeoKGs, S2 Geometry offers significant advantages \citep{zalewski2021semantic}. Distance and area calculations in WGS84, the most commonly used coordinate reference system (CRS) in GIS, require geodesic computations on an ellipsoid, which can be computationally expensive for large datasets. In contrast, S2 Geometry simplifies these calculations by approximating distances and areas on a unit sphere. Although this results in slightly less accurate measurements, it significantly improves computational efficiency. The regular hierarchical structure of S2 ensures that point-in-cell relationships are preserved across multiple resolutions, enabling reliable and scalable qualitative spatial querying. Additionally, the near-equal area of S2 cells is critical for statistical aggregations, such as approximating quantities within a cell by summing over all child cells.


\vspace{10pt}

\vspace{7pt}

The core S2 Geometry library, licensed under the open-source Apache License 2.0, is written in C++ \citep{veach2017s2}. Wrappers for this library are available in Python, Java, JavaScript, R, and other languages at varying levels of maturity. The primary purpose of the S2 library is to provide operations for computational geometry and spatial indexing on the sphere. The S2 library supports the following geospatial functionalities:

\begin{itemize}
    \item Representations of angles, intervals, and geometric shapes (e.g., points, polylines, polygons) both on the unit sphere and in latitude--longitude space.

    \item Constructive operations (e.g., union, difference, buffer), boolean predicates for testing topology relationships (e.g., containment, intersection), and mathematical predicates for evaluating spatial relationships (e.g., distance within a threshold, point-in-region) for arbitrary geometric objects.

    \item Algorithms for measuring areas, calculating distances, computing centroids, simplifying geometry, and finding nearby objects.

    \item Functions for cell navigation (retrieving parent, children, and neighboring cells), and traversing cells at the same level along the Hilbert curve.

    \item Support for spatial indexing, including conversion between geographic coordinates and S2 cells, and the ability to approximate arbitrary regions as unions of S2 cells (see Figure~\ref{fig:regionCoverer}).
\end{itemize}

\section{Methodology}\label{sec:methodology}

This section outlines the technical aspects of integrating and utilizing S2 within KWG. It provides an overview of how S2 geometry is modeled and ingested into KWG, followed by a discussion of the general technical approach for implementing various quantization methods. These methods will be further detailed in Section~\ref{sec:implementation}.

\subsection{Conceptual Schema for the S2 Geometry in KnowWhereGraph}

GeoSPARQL 1.1 \citep{car2022geosparql} introduces the $\texttt{geo}$:$\texttt{asDGGS}$ property to link a DGGS cell to its cell ID serialization in the $\texttt{geo}$:$\texttt{dggsLiteral}$ datatype. However, GeoSPARQL does not natively interpret DGGS literals, necessitating external implementations for handling them. Currently, GeoSPARQL-DGGS \citep{geosparql-dggs} is the only library available that leverages RDFlib to support GeoSPARQL's topological functions on DGGS literals, but it is limited to the rHEALPix DGGS. Given this limitation, we chose to represent S2 cells as vector geometries (with cell vertices forming a polygon) using the WKT specification. This approach ensures compatibility with existing geospatial standards and tools, providing a robust solution for integrating S2 cells within GeoSPARQL-extending frameworks such as KWG.

\vspace{10pt}
\begin{figure}[h]
    \centering
    \includegraphics[width=0.45\linewidth]{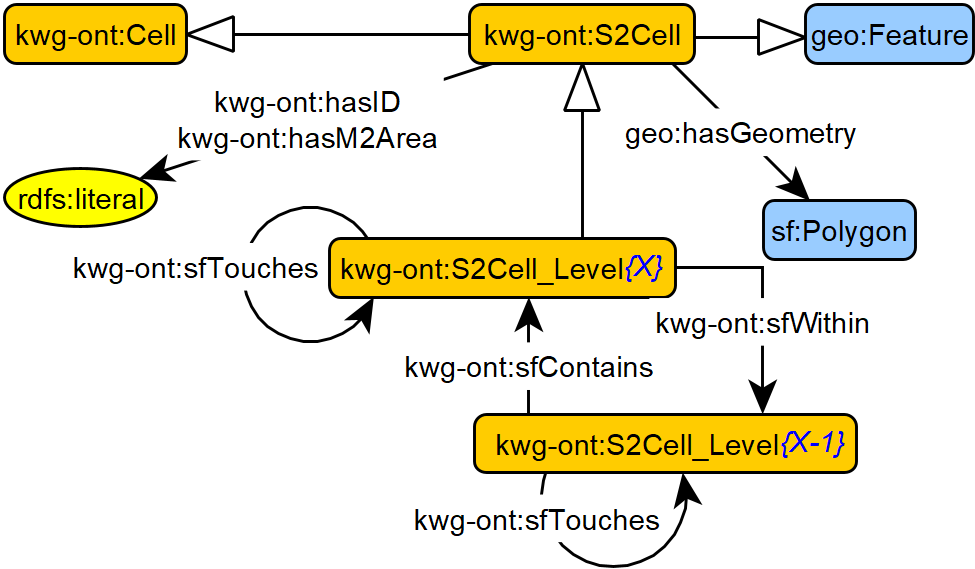}
    \caption{\normalsize Schema diagram illustrating the extension of GeoSPARQL concepts to model S2 data.}
    \label{fig:s2-schema}
\end{figure}
\vspace{7pt}

The schema diagram in Figure~\ref{fig:s2-schema} represents the conceptual modeling of the S2 Geometry in KWG. The primary class, $\texttt{kwg-ont:S2Cell}$, is defined as a subclass of $\texttt{geo}$:$\texttt{Feature}$ and $\texttt{kwg-ont:Cell}$, generalizing cells in a DGG \citep{shimizu2021pattern}. The geometry of each S2 cell is a simple feature polygon. The more detailed geometric representation of S2 cells is adopted in KWG primarily for visualization needs, but if visualization and runtime topological querying are not a priority this geometry can be abstracted to a point. $\texttt{kwg-ont:S2Cell}$ is further specialized into subclasses denoted as $\texttt{kwg-ont:S2Cell}$\_$\texttt{Level\{X\}}$, where $\texttt{X}$ indicates various levels in the S2 hierarchy. Cells are structurally connected both within and between levels via DE-9IM relationships from the KWG ontology (Figure~\ref{fig:kwg-core}). Hierarchical \emph{parent--child} relationships between S2 cells are represented by $\texttt{kwg-ont:sfContains}$ and $\texttt{kwg-ont:sfWithin}$, while \emph{neighbor} relationships are expressed using $\texttt{kwg-ont:sfTouches}$. Hierarchical connectivity is explicitly precomputed up to one level, while multi-level connectivity is inferred via the transitive properties defined on $\texttt{kwg-ont:sfContains}$ and $\texttt{kwg-ont:sfWithin}$. In a reasoning-enabled GraphDB repository, these inferred relationships are pre-materialized through forward-chaining during the data loading process. Additionally, each S2 cell is annotated with its identifier and area using the data properties $\texttt{kwg-ont:hasID}$ and $\texttt{kwg-ont:hasM2Area}$, while the cell resolution is implicit in the class name.

\begin{figure}
    \centering
    \includegraphics[width=0.8\linewidth]{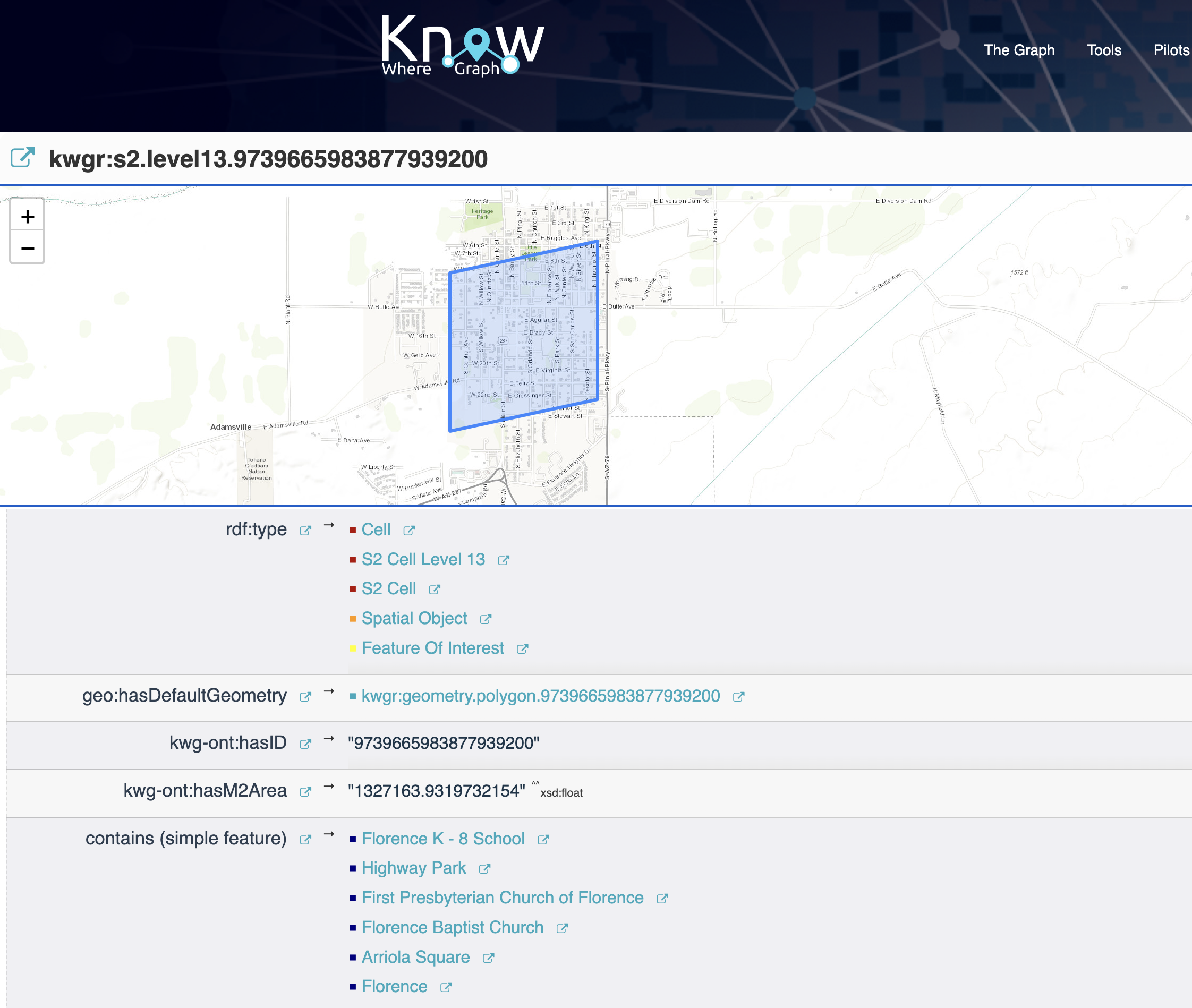}
    \caption{\normalsize S2 cell with cell ID 9739665983877939200 visualized within KWG's phuzzy dereferencing application.}
    \label{fig:s2-phuzzy}
\end{figure}

The \pyth{s2sphere}\footnote{\url{http://s2sphere.readthedocs.io/}} python implementation of the S2 Geometry library is used in generating the S2-RDF graph, a subgraph in KWG. Figure~\ref{fig:s2-phuzzy} illustrates one S2 cell from the graph displayed on a map via KWG's dereferencing application. The snapshot also displays the attributes and linkages between the cell and other features in the graph.

\subsection{Challenges}

Using S2 cells as polygon geometry representations in WKT serialization introduces certain ambiguities. The discrepancies between edges in planar and spherical spaces can result in indexing challenges and visualization artifacts, which are discussed in this section.

\begin{figure}[h]
    \centering
    \begin{subfigure}[t]{0.4\textwidth}
        \includegraphics[width=\textwidth]{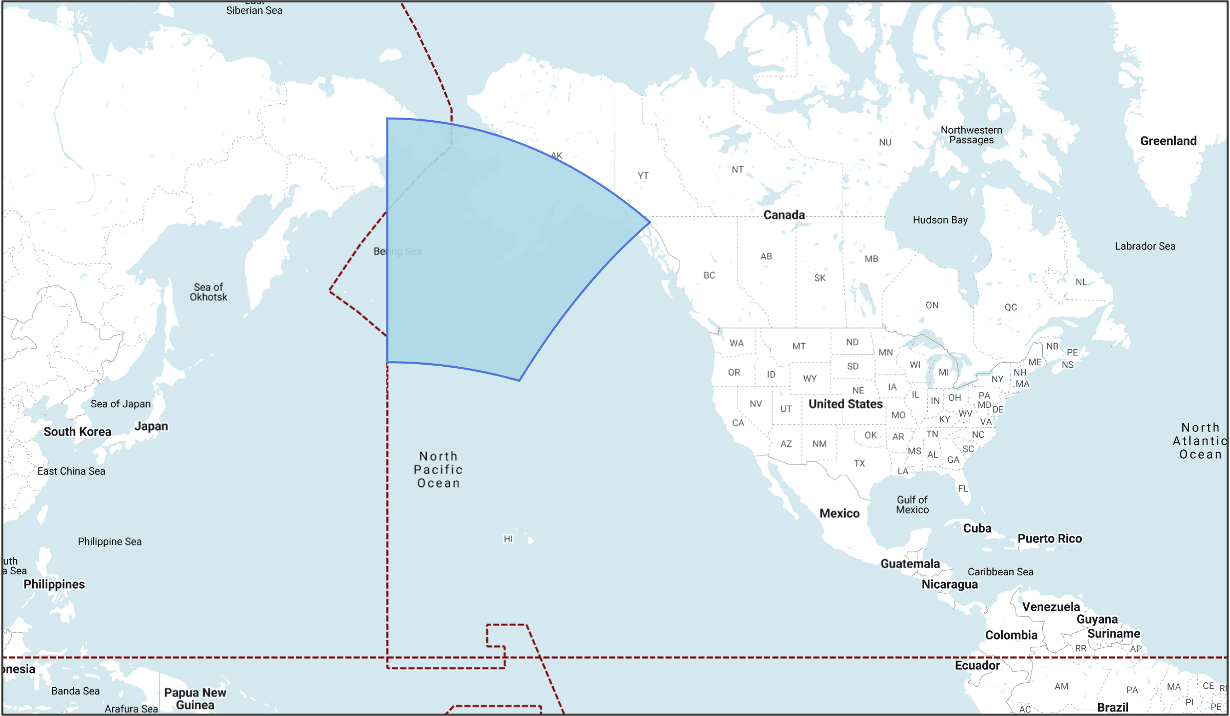}
        \caption{With wrapping (using BigQuery Geo Viz)}
        \label{fig:s2.1}
    \end{subfigure}
    ~\qquad
    \begin{subfigure}[t]{0.55\textwidth}
        \includegraphics[width=\textwidth]{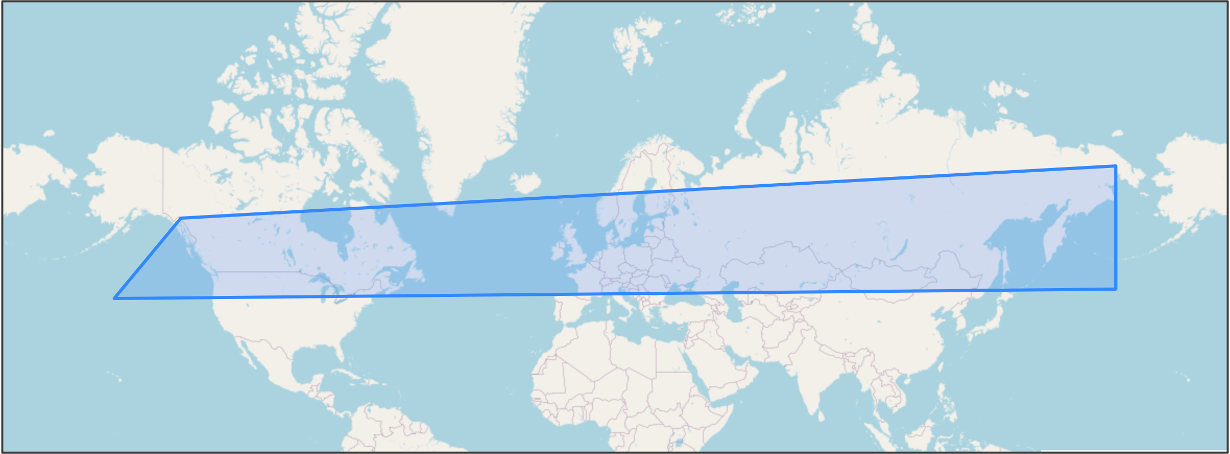}
        \caption{Without wrapping (using wktmap.com)}
        \label{fig:s2.2}
    \end{subfigure}
    \caption{S2 cell with cell ID 9007199254740992000. Its WKT serialization is \pyth{''POLYGON((180 67.3801, 180 45, -157.3801 42.7093, -135 59.4910, 180 67.3801))''}$\char`\^\char`\^$ \pyth{geo:wktLiteral}.}\label{fig:s2}
\end{figure}

\vspace{7pt}
\emph{\textbf{Indexing:}} The default CRS for WKT literals is WGS84. Geographic coordinates in WGS84 span from $180^{\circ}$ to $180^{\circ}$ longitude and -$90^{\circ}$ to $90^{\circ}$ degrees latitude, but some S2 cells span the antimeridian ($\pm180^{\circ}$) and the poles. Since indexing in GraphDB uses WGS84, indexing issues with cells that cross the antimeridian or the poles because they are not normalized (either through wrapping or splitting) rendering them topologically invalid. Figure~\ref{fig:s2.1} illustrates an example of an S2 cell crossing the antimeridian that cannot be indexed in the graph. This issue occurs when the longitude difference between two consecutive points in the path is >=$180^{\circ}$. In spatial databases like GraphDB, this ambiguity must be resolved to ensure accurate representation and querying. Possible solutions include splitting S2 cells at the antimeridian or representing them as points.

\vspace{10pt}

\begin{figure}[h]
    \centering
    \begin{subfigure}[t]{0.25\textwidth}
        \includegraphics[width=\textwidth]{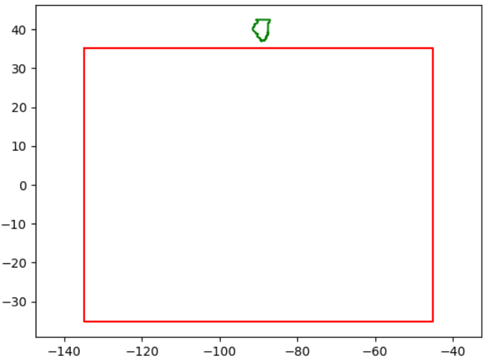}
        \caption{}
        \label{fig:illinois.1}
    \end{subfigure}
    ~\qquad
    ~\qquad
    \begin{subfigure}[t]{0.25\textwidth}
        \includegraphics[width=\textwidth]{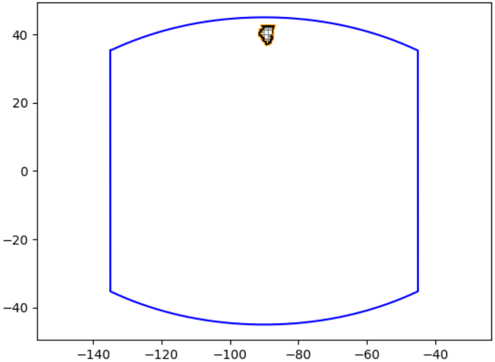}
        \caption{}
        \label{fig:illinois.2}
    \end{subfigure}
    \caption{Visualization artifacts when geodesic edges of S2 cells are considered as rhumb lines, or paths of constant bearing relative to north. (a) Illinois is not inside S2 cell, level 0, hex 9, which has straight-edge boundaries. (b) Illinois is within S2 cell, level 0, with cell ID hex 9, showing geodesic boundaries. Hex refers to the hexadecimal representation of the S2 cell ID.}\label{fig:error-visualization}
\end{figure}

\emph{\textbf{Visualization Artifacts:}} When S2 cells are visualized on maps with a Mercator projection, as in most web maps, their true shape and size on the sphere may not be accurately reflected, with more pronounced distortion at higher latitudes. S2 uses geodesic edges, meaning that the edges of the cells follow the shortest path on the spherical Earth approximation. On a Mercator projection, geodesic paths in the east--west direction, except at the Equator, appear as curves bulging towards the poles.  Consequently, the ``horizontal'' edges of S2 cells appear curved towards the poles (Figure~\ref{fig:illinois.2}). When S2 cells are represented by straight lines connecting their four corner vertices in a Mercator projection, their east--west edges are incorrectly rendered as rhumb lines, or paths that follow a constant bearing relative to north, like lines of latitude (Figure~\ref{fig:illinois.1}). This shape distortion can lead to potential misinterpretations of the spatial data. For example, points that are topologically within an S2 cell may appear to lie outside the cell, and vice versa (an example using the state of Illinois is compared in Figure~\ref{fig:error-visualization}). Additionally, S2 quadrilaterals do not tessellate perfectly on planar maps, leading to gaps or overlaps between cells. Figure \ref{fig:s2.2} denotes another visualization artifact, where an S2 cell that crosses the antimeridian appears stretched across the world when represented on a flat map that uses Mercator map projection.
\vspace{10pt}

\subsection{Technical Strategy for Data Quantization in KnowWhereGraph}

S2 in KWG facilitates efficient data storage and retrieval through different data quantization strategies. As previously mentioned, indexing and query operations provided by the S2 Geometry are not natively available within GraphDB. Quantization is adapted in the KWG environment by way of view materialization \citep{ibragimov2016optimizing}, where triples --- spatial, temporal, thematic, or their combinations --- are pre-computed using the \pyth{s2sphere} library and persisted within the graph. These materialized views enhance the scalability of query processing, particularly for complex spatial analytic operations, by circumventing direct access to geometric data (e.g., WKT literals) within the graph. We employ tailored ontology patterns \citep{falbo2013ontology} to guide the selection of views for materialization \citep{lan2022answering}. These patterns, detailed in Section~\ref{sec:implementation}, delineate different ways for quantizing data to S2 cells. Nodes in a pattern depict frequently accessed relationships (based on use-case needs) or spatial joins earmarked for materialization, thereby mitigating query latency.

Precomputation of triples also facilitates data conflation by consolidating vector, raster, and other geographically themed datasets at the intersection of the S2 grid. This process enriches the graph, improving the efficacy of graph-based algorithms such as similarity detection and clustering, and fostering serendipitous discoveries across interconnected data entities. Unlike the space costs associated with relational databases, materializing views in a graph database also allows for data compression, which would be impossible if data were ingested with their original structure and precision. Whether for enrichment or compression, the choice of pattern depends on a nuanced consideration of factors such as on-the-fly computation overheads, graph size, and graph maintenance.


\section{Implementing Data Quantization on S2 in KnowWhereGraph}\label{sec:implementation}

In this section, we elaborate on the two quantization strategies adopted in KWG, the ontology patterns adopted for materializing views, and their specific benefits.

\subsection{Topological Enrichment of Vector Data}\label{sec:enrichment}
Research on computational time and space trade-offs in GeoKGs has demonstrated that precomputation of topological relations can yield significant performance benefits \citep{regalia2016volt, regalia2019computing}. Traditionally, places (represented as \texttt{kwg-ont:Region} nodes in KWG) have been central for topologically linking other spatial features (e.g., \texttt{kwg-ont:Hazard} and \texttt{kwg-ont:RoadSegment}) within GeoKGs \citep{regalia2019computing}. However, place boundaries are inherently dynamic, meaning their topological relationships with intersecting features can change over time. For instance, before 2011, South Sudan was spatially \emph{within} Sudan, but after gaining independence, the relationship changed to a border-sharing or \emph{touches} relationship. In contrast, S2 cell boundaries are fixed and immutable, providing a stable spatial framework for linking features and supporting multi-resolution and hierarchical spatial reasoning \citep{timpf1997using}.

\begin{figure}[h]
  \begin{subfigure}[b]{0.36\textwidth}
    \vspace{0pt}
    \includegraphics[width=\textwidth]{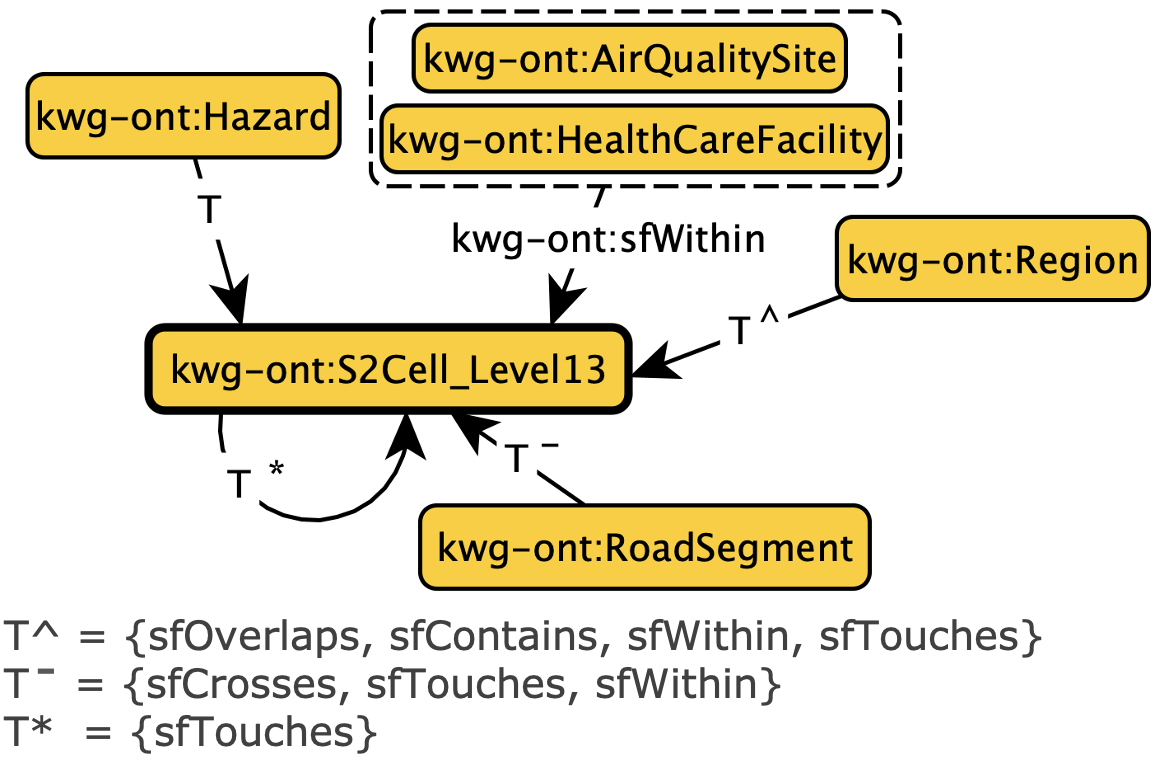}
     \caption{}
        \label{fig:topo-pattern}
  \end{subfigure} 
  \begin{subfigure}[b]{0.6\textwidth}
{\renewcommand{\arraystretch}{1.2}%
\begin{tabular}{|l|lll|}
\hline
\multicolumn{1}{|c|}{\textbf{\begin{tabular}[c]{@{}c@{}}kwg-ont:spatialRelation\\ (T )\end{tabular}}} & \multicolumn{1}{c|}{\textbf{\begin{tabular}[c]{@{}c@{}}geometry\\ combination\end{tabular}}} & \multicolumn{1}{c|}{\textbf{property}} & \multicolumn{1}{c|}{\textbf{\begin{tabular}[c]{@{}c@{}}precomputed\\ vs. inferred\end{tabular}}} \\ \hline
kwg-ont:sfTouches & \multicolumn{1}{l|}{all except P/P} & \multicolumn{1}{l|}{symmetric} & precomputed \\ \hline
kwg-ont:sfOverlaps & \multicolumn{1}{l|}{P/P, L/L, A/A} & \multicolumn{1}{l|}{symmetric} & precomputed \\ \hline
kwg-ont:sfContains & \multicolumn{1}{l|}{all} & \multicolumn{1}{l|}{inverse} & precomputed \\ \hline
kwg-ont:sfWithin & \multicolumn{1}{l|}{all} & \multicolumn{1}{l|}{transitive} & inferred \\ \hline
kwg-ont:sfEquals & \multicolumn{3}{l|}{\begin{tabular}[c]{@{}l@{}}Not precomputed or inferred, but relations are\\ included in the ontology for completeness\end{tabular}} \\ \hline
kwg-ont:sfDisjoint & \multicolumn{3}{l|}{\begin{tabular}[c]{@{}l@{}}Not precomputed or inferred, but relations are\\ included in the ontology for completeness\end{tabular}} \\ \hline
\end{tabular}%
}
    \caption{}
        \label{fig:relation}
  \end{subfigure}
  \caption{\textbf{(a)} Ontology pattern for topologically enriching vector geographic entities. \textbf{(b)} Table showing the properties asserted on topological relations in KWG for various geometry type combinations (P-Point, L-Curve, A-Surface) and the corresponding A-Box triples that are materialized.}
\end{figure}

The S2 grid at level 13 is selected as the reference layer, offering an appropriate level of detail (\char`~$1.27$ $km^2$ per cell) for topologically interlinking KWG data. From this point forward, cells in this layer will be referred to as S2 reference cells. \emph{Topological enrichment of vector data} is achieved by precomputing and materializing topological relations between entities with explicit geometries and S2 reference cells in the graph, creating a dense network of interconnected spatial objects. The spatial relations utilized are a subset of topological relations from $\mathcal{T}$ (see Section~\ref{sec:kwg}), which are sub-properties of $\texttt{kwg}$-$\texttt{ont}$:$\texttt{spatialRelation}$. Figure~\ref{fig:topo-pattern} illustrates the ontology pattern for this view materialization. Certain datasets ingested into KWG (e.g., public health departments, and pharmacies) provide addresses or other location-based information, such as ZIP codes, which are geocoded into point coordinates for topological linking to S2 cells.

Figure~\ref{fig:relation} represents the OWL properties \citep{hitzler2009owl} defined over the topological relations in $\mathcal{T}$. We precompute most of the implicit triples induced by properties to reduce graph initialization costs (\emph{total materialization} at load time). We leave the triples resulting from the transitive property (on $\texttt{kwg}$-$\texttt{ont}$:$\texttt{sfWithin}$) to be entailed from forward chaining during graph initialization.

\vspace{10pt}

\vspace{7pt}

This enrichment, although it may significantly increase the number of triples in the graph, makes feature geometries obsolete, thereby enhancing querying and data processing speed. Table~\ref{tab:query-times} shows the spatial query speedups achieved through topological enrichment.

\vspace{8pt}

\begin{table}[h]
\resizebox{\textwidth}{!}{%
\begin{tabular}{|c|l|c|c|c|c|}
\hline
\multicolumn{1}{|l|}{\textbf{Query type}} & \multicolumn{1}{c|}{\textbf{Query details}} & \textbf{\begin{tabular}[c]{@{}c@{}}No. results retrieved\\ (via S2)\end{tabular}} & \textbf{\begin{tabular}[c]{@{}c@{}}Query time\\ (via S2)\end{tabular}} & \multicolumn{1}{r|}{\textbf{\begin{tabular}[c]{@{}r@{}}No. results retrieved\\ (via GeoSPARQL)\end{tabular}}} & \textbf{\begin{tabular}[c]{@{}c@{}}Query time\\ (via GeoSPARQL)\end{tabular}} \\ \hline
P/A & hospitals in Santa Barbara {[}Q1{]} & 22 & 0.3 secs & 23 & 30 secs \\ \hline
L/A & roads intersecting Santa Barbara {[}Q2{]} & 1747 & 5.7 secs & - & Timeout \\ \hline
A/A & statistical areas overlapping Santa Barbara {[}Q3{]} & 4 & 1.9 secs & 4 & 5.8 secs \\ \hline
\end{tabular}%
}
\caption{\normalsize Querying the graph using S2 (with topological enrichment) versus using spatial functions from GeoSPARQL. Queries are available in the following Github repository: \url{https://github.com/shirlysteph/s2-kwg/tree/main/demos/sample-queries}}
\label{tab:query-times}
\end{table}
\vspace{7pt}

\noindent \textbf{\emph{A. Optimized Topological Enrichment: }}

Classic topological enrichment uses S2 cells at level 13, but applying this resolution to large polygons, such as drought or smoke plume areas, can result in inefficiencies due to the massive data volumes involved. For example, a temporally sliced multi-polygon representing drought could cover up to 6.8 million square kilometers of the contiguous United States. To improve efficiency, a discrete and compressed enrichment framework \citep{zalewski2021semantic} is employed, combining DE-9IM with hierarchical S2 cell nesting. This method links large polygons to the largest S2 cells at coarser resolutions, leveraging the transitivity of within/containment relations in the spatial ontology to infer triples describing spatial relationships between the region and S2 reference cells. This approach reduces data volume and speeds up computations by limiting the number of cells processed in each query.

\vspace{10pt}
\begin{figure}[h]
    \centering
    \begin{subfigure}[t]{0.29\textwidth}
        \includegraphics[width=\textwidth]{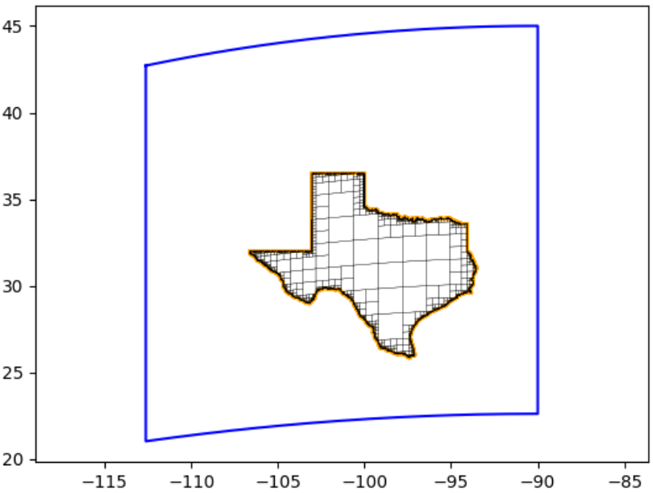}
        \caption{}
        \label{fig:cropland}
    \end{subfigure}
    ~\qquad
    \begin{subfigure}[t]{0.6\textwidth}
        \includegraphics[width=\textwidth]{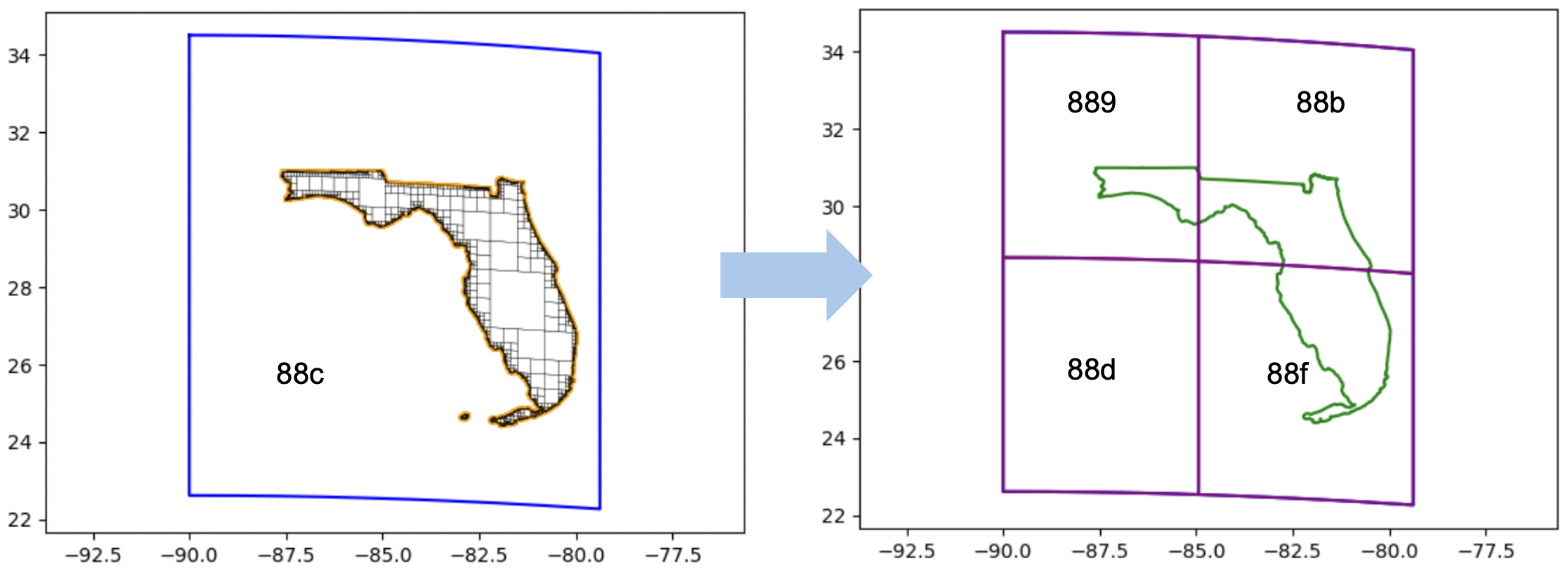}
        \caption{}
        \label{fig:soil}
    \end{subfigure}
    \caption{\normalsize Examples of optimized topological enrichment: (a) Texas is covered with S2 cells at various levels for efficient representation. (b) Florida is similarly covered with S2 cells at different levels and is contained within one S2 cell at level 3 (hex 88c), while overlapping four S2 cells at level 4 (hex 889, 88b, 88d, 88f).}\label{fig:optimized-integration}
\end{figure}

\vspace{7pt}
\noindent \textbf{\emph{B. Computational Method for Boolean Topological Relationships: }}
The S2 Geometry library provides classes for representing shapes (points, curves, and regions) on the Earth's surface. These shapes are defined using points and geodesic curves, which presents challenges when working with vector data serialized in the WGS84 format, as these data do not inherently conform to geodesic boundaries. To address this incompatibility, we developed a custom algorithm to compute topological relationships (based on the DE-9IM) between the S2 reference cells and geometric objects in KWG. This algorithm leverages the \pyth{s2sphere} python wrapper for the library alongside \pyth{Shapely}, which handles planar geometric objects. Key methods from the S2 library's \texttt{S2RegionCoverer} class, such as \texttt{GetCovering} and \texttt{GetInteriorCovering}, were instrumental in generating \texttt{S2CellUnion} objects that either cover or are contained within the specified region. The technical steps of the algorithm are as follows:

\vspace{-10pt}
\begin{enumerate}
    \item  \textbf{Geometry approximation:} Each WGS84 geometry ([multi-]point, [multi-]line, [multi-]polygon) is approximated as its analogous S2 object(s) on the unit sphere. In the case of polygons, the boundary is segmented by adding vertices such that the distance between any two consecutive vertices does not exceed a specified tolerance. This ensures that linear edges are replaced by geodesic curves that closely approximate the original planar shapes.
    \item \textbf{Creating coverings:} The \pyth{S2RegionCoverer} is then used to approximate the S2 object as unions of S2 reference cells, known as cell coverings. This includes ordinary coverings (cells at various levels that cover the geometry), homogeneous coverings (cells that cover the geometry at a specified level), and interior coverings (cells at various levels strictly within the geometry).  
    \item \textbf{Topological relationship computation:} For point data, each point is directly mapped to an \pyth{S2Point} and associated with an \pyth{S2CellID} at the maximum resolution level. For line and polygon data, topological relationships are computed by identifying overlapping or contained cells using the coverings generated. For example, to identify the cells at a specified level that are strictly within the given shape, an interior covering is used. Overlaps are determined by the difference between homogeneous and interior coverings.

\end{enumerate}

\subsection{Grid-Based Data Discretization}

\begin{figure}[h]
    \centering
    \includegraphics[width=0.55\linewidth]{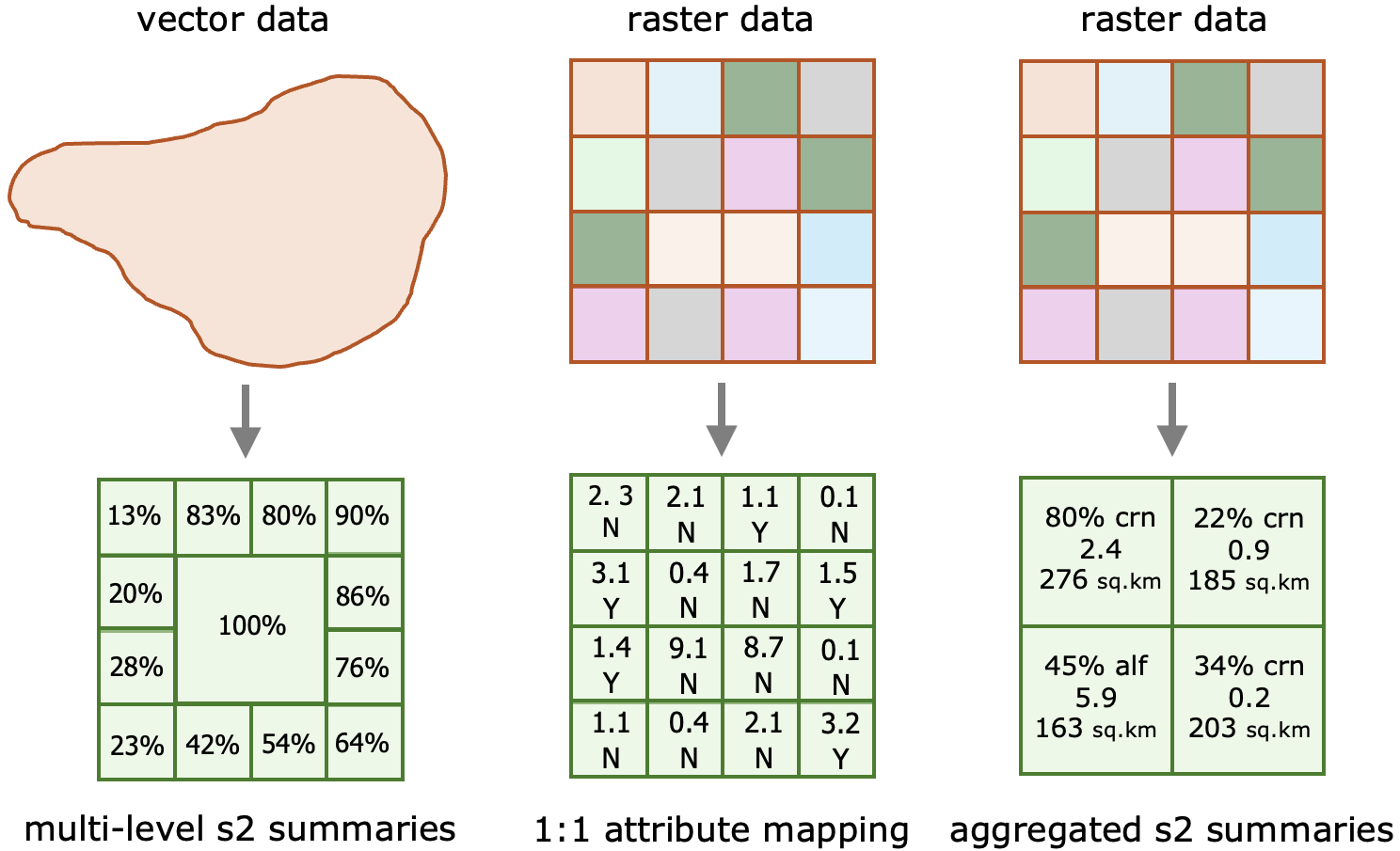}
    \caption{{\normalsize Examples of discretizing raster and vector data on the S2 grid.}}
    \label{fig:quantization}
\end{figure}

Data discretization involves transforming non-DGGS spatial data (vector or raster) through different transformation strategies (statistical aggregation, decomposition, spatial overlay, etc.) to represent them as S2 cell--based gridded data for flexible representation and analysis while maintaining spatial relationships and scalability across different resolutions. Cells are indicated as the $\texttt{sosa:FeatureOfInterest}$ \citep{janowicz2019sosa} to which various data values are mapped, via object properties or data properties. These data values can be 1:1 grid mapping (for raster cells), aggregated summaries, or multi-level decomposition --- see examples in Figure~\ref{fig:quantization}. Quantized values can be quantitative (e.g., area covered by a smoke plume) or qualitative details (e.g., categorical data related to soil porosity).


Vector data in KWG spans a wide range of scales, from large-scale features like smoke plumes covering extensive areas to highly localized data such as wildfire events represented as points or small polygons. Although wildfires are a primary source of smoke plumes, their spatial and temporal distributions provide different insights. Wildfire distribution can reveal persistent hotspots and frequent fire activity, while smoke plume distribution is essential for assessing correlations with health impacts. Analyzing both datasets together enhances understanding of regional air quality and supports the development of smoke forecast models. However, comparing these disparate features without appropriate scaling and context can lead to spatial mis-contextualization, where localized patterns or anomalies may be obscured, potentially resulting in incorrect interpretations of their interrelationships. Identifying the optimal scale for analysis --- whether for correlation or hotspot detection --- is not always straightforward, underscoring the need for data solutions that support multi-scale analysis \citep{manley2021scale}. Discretizing large-scale vector features by decomposing them into varying resolutions of S2 cells can help reveal critical relationships and patterns that may be obscured at other scales.

\vspace{10pt}

Non-DGGS data can also be aggregated on the S2 grid as summaries (e.g., kernel density, spatial statistics, histograms, frequency transformations) to offer insights into aggregated spatial patterns without necessitating the retention of detailed individual records. Aggregation can occur at different levels of spatial granularity. For instance, real-time fire data can be aggregated at fine spatial scales (e.g., neighborhood or street levels) to facilitate targeted response efforts and resource allocation within localized areas. Aggregating the same fire data at higher spatial levels (e.g., city or county levels) helps in understanding broader phenomena such as air pollution or regional fire trends. Summaries computed at a lower level can be aggregated upward through the S2 hierarchy to compute values for larger spatial regions. Summaries can also be decomposed downward within the hierarchy, provided the feature in consideration is either $\texttt{kwg-ont:sfEqual}$ or $\texttt{kwg-ont:sfWithin}$ the S2 cell at the top level. This hierarchical approach supports adaptive analysis, where users can dynamically adjust the level of decomposition to focus on specific spatial features or patterns of interest.

Both data decomposition and aggregation through discretization facilitate multi-scale analysis for both spatial and non-spatial dimensions of the data. The specific choice of method and selected grid resolution should reflect both the purpose and scale of the analysis. The hierarchical grid structure of S2 supports seamless transitions between different scales --- from local to global --- allowing for detailed representation and the discovery of patterns that may not be evident at a single scale. Furthermore, this quantization method improves data management by supporting feature simplification. It also supports advanced geovisualization techniques, allowing for dynamic scaling and context-aware rendering. This ensures that large-scale patterns and localized details are preserved and accurately represented, enhancing the ability to uncover meaningful spatial relationships and insights.

Here we describe two examples of data discretization in KWG: one for a vector dataset and another for a raster dataset.


\subsubsection{Vector Data to S2 Cell Discretization}
The Soil Survey Geographic Database\footnote{\url{https://websoilsurvey.nrcs.usda.gov/app/WebSoilSurvey.aspx}} (SSURGO) contains detailed information on hundreds of soil attributes,  including chemical and physical properties, as well as their derived interpretations. The dataset comprises over 36 million soil map units distributed across more than 70 shapefiles, with varying scales depending on the region. For example, states like Nevada and Utah have large polygons mapped at a scale of 1:63,360, while agriculturally intensive states like Indiana and Iowa have finer polygons at a scale of 1:12,000. Due to its size and complexity, SSURGO data can be impractical for use in integrated analytics environments. In KWG, this dataset is decomposed onto the S2 reference cells. Figure~\ref{fig:soil} demonstrates the ontology pattern employed. Overlap areas between individual soil polygons and an S2 reference cell (measured in square kilometers) are precomputed and materialized as instances of the $\texttt{kwg-ont:SoilMapUnitS2OverlapObservation}$ class, with the S2 cell serving as the feature of interest. The geometries of the original soil map units, represented by the $\texttt{kwg-ont:SoilMapUnit}$ class, are also included in the graph along with the associated soil attributes, modeled as $\texttt{kwg-ont:SoilMapUnitObservableProperty}$. These geometries enable more accurate visualizations, as a single S2 cell can contain disconnected geometries from a soil multipolygon. Through the composition of relationships, $\texttt{kwg-ont:overlapObservedWith}$ $\circ$ $\texttt{sosa:isFeatureOfInterestOf}$, users can efficiently explore all soil properties linked to an S2 cell without the need to directly access soil polygon geometries. By discretizing soil data over a high-resolution S2 grid, the level of detail and accuracy is enhanced, allowing for integration with other thematic datasets such as landforms, floodplains, crop types, and high-resolution digital elevation models.
\vspace{8pt}

\begin{figure}[h]
    \centering
    \includegraphics[width=0.8\linewidth]{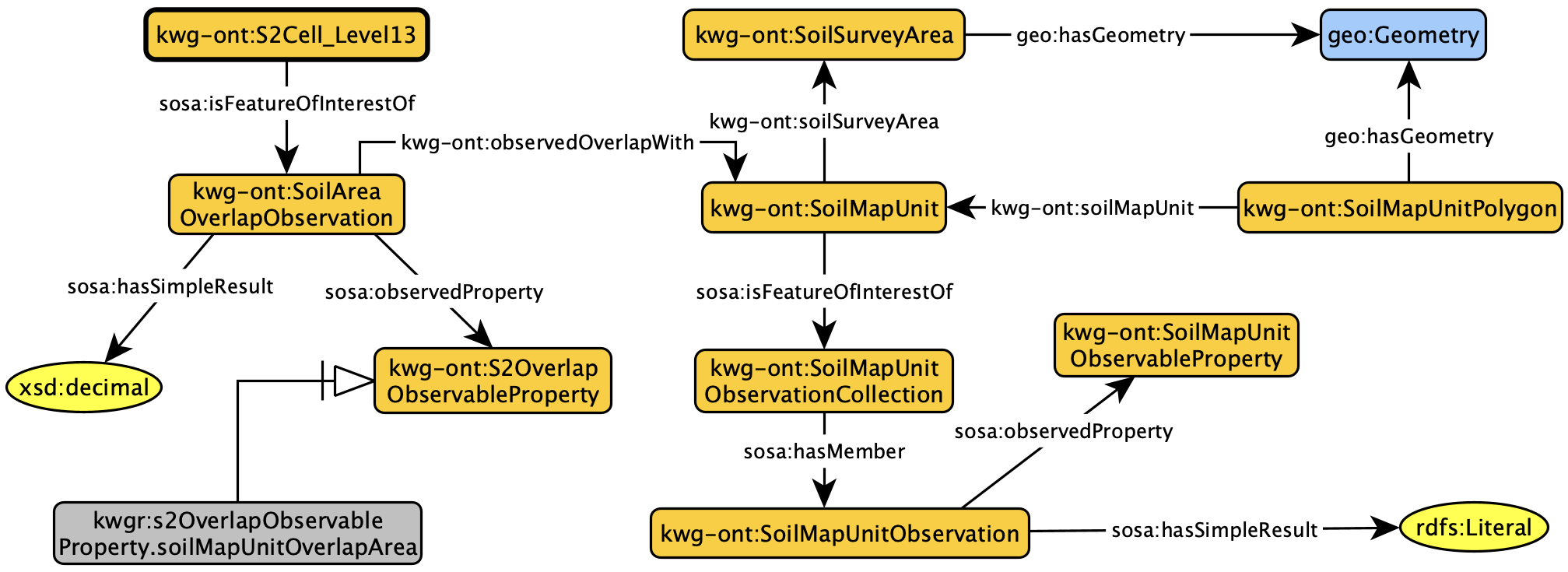}
    \caption{\normalsize Ontology pattern for discretizing vector soil data from SSURGO to S2 reference cells.}
    \label{fig:soil}
\end{figure}

To discretize SSURGO data, we build upon the method for computing topological relationships between vector geometries and S2 cells, as outlined in Section~\ref{sec:enrichment}. Vector geometries of $\texttt{kwg-ont:SoilMapUnitPolygon}$ geometries are approximated as \texttt{S2Polygon} objects, and their cell coverings are generated using S2 reference cells. The overlaps between the cells in the covering and the \texttt{S2Polygon} are computed using the \texttt{S2Polygon.GetOverlapFractions} method, which returns the fractional overlap of each cell relative to the polygon. This fraction is then multiplied by the cell's area to compute the actual overlapping area on the Earth's surface. This quantifiable measure allows for a more detailed understanding of soil properties associated with reference S2 cells.


\subsubsection{Raster Data to S2 Cell Discretization}
The USDA National Agricultural Statistics Service Cropland Data Layer\footnote{\url{https://croplandcros.scinet.usda.gov/}} (CDL) is a raster-based land cover dataset produced annually using satellite imagery and extensive agricultural ground reference data. The dataset has a ground resolution ranging from 30 to 56 meters, depending on the state and year. It tracks changes in forest extent and height, cropland, built-up areas, surface water, perennial snow and ice, and over 134 distinct crop types, making it highly valuable for monitoring agricultural land use in KWG. To integrate this data into KWG, the pixel-based raster information is transformed into data values representing the areal extent of each crop type within an S2 reference cell. Figure~\ref{fig:cropland} illustrates the ontology pattern for this materialization in the graph. The areal percentage of each crop type within a reference cell, modeled as $\texttt{kwg-ont:CroplandObservableProperty}$, is computed and materialized as a temporally scoped observation instance of $\texttt{kwg-ont:CroplandS2OverlapObservation}$. To ensure data accuracy, we verify that the sum of all land cover observations within each cell, for a given timestamp, equals 100\%. This discretization process reduces the large volume of raster CDL data and represents it in the graph, facilitating more efficient and scalable computations. As a result, queries are attribute-based rather than spatial, improving both performance and manageability.
\vspace{8pt}

\begin{figure}[h]
    \centering
   \includegraphics[width=0.7\linewidth]{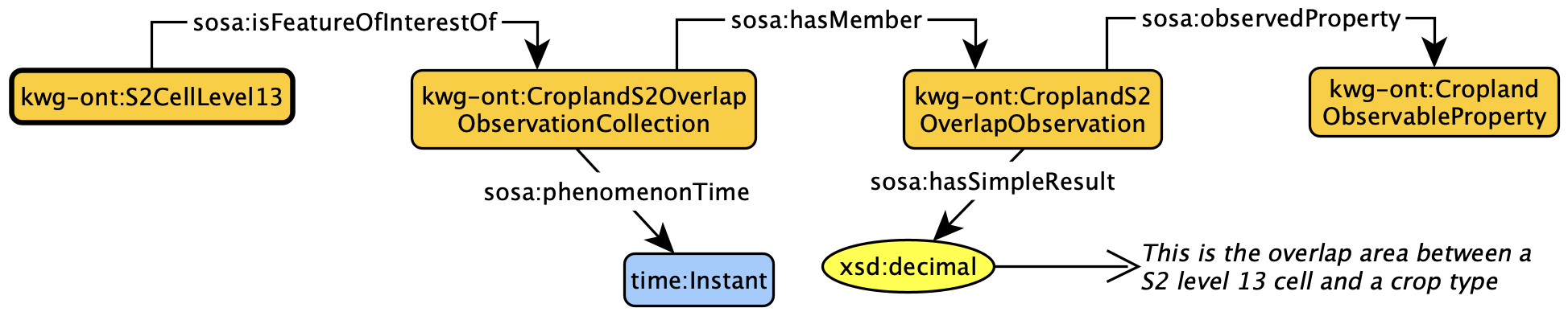}
    \caption{\normalsize Ontology pattern for discretizing raster cropland data from USDA CDL to S2 reference cells.}
    \label{fig:cropland}
\end{figure}

The discretization of cropland data with S2 reference cells involves transforming raster representations into S2-compatible geometries to analyze spatial relationships more efficiently. Raster data is typically associated with a specific CRS, which must be transformed to align with the spherical geometry of the S2 library before generating S2 cell coverings and overlap calculations. The process begins by extracting the spatial bounds of the raster dataset from its metadata and defining the geographical coordinates of its corners. These bounds are then converted into an \texttt{S2Polygon}, which represents the geographical area covered by the raster on the spherical Earth. This conversion ensures the raster’s spatial footprint is accurately mapped onto the spherical surface for further analysis. Once the \texttt{S2Polygon} is established, the \texttt{S2RegionCoverer} method is used to cover the polygon with S2 reference cells. For each S2 cell in the covering, the raster values within the cell are aggregated to compute statistical summaries, such as the mean, sum, or other metrics relevant to the data. These statistical summaries provide quantifiable measures of the raster data's distribution within each S2 cell.




\section{Analytical Capabilities in KnowWhereGraph With S2 Integration}\label{sec:usage}

This section presents four use cases that demonstrate how the two quantization strategies enhance the analytical pipeline in KWG by optimizing storage and computation, while improving spatial query efficiency over heterogeneous datasets.

\subsection{Conflating S2 Cells with Thematic Observation Data via Object Property Paths}

\begin{figure}
    \centering
    \includegraphics[width=0.85\linewidth]{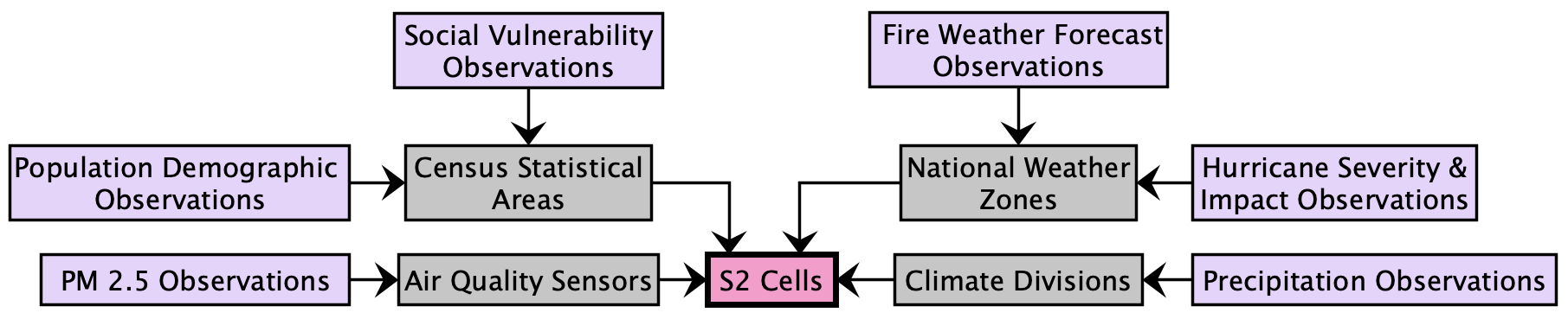}
    \caption{This figure illustrates the contextual enrichment of S2 cells with various observational data via geographic entities in KWG. Purple boxes represent observation data, while grey boxes denote geographic features with explicit geometries. Arrows indicate the linkages connecting observations to features and, in turn, to the corresponding S2 cells.}
    \label{fig:conflation}
\end{figure}

\begin{figure}[h]
  \begin{subfigure}[b]{0.44\textwidth}
    \vspace{0pt}
    \includegraphics[width=\textwidth]{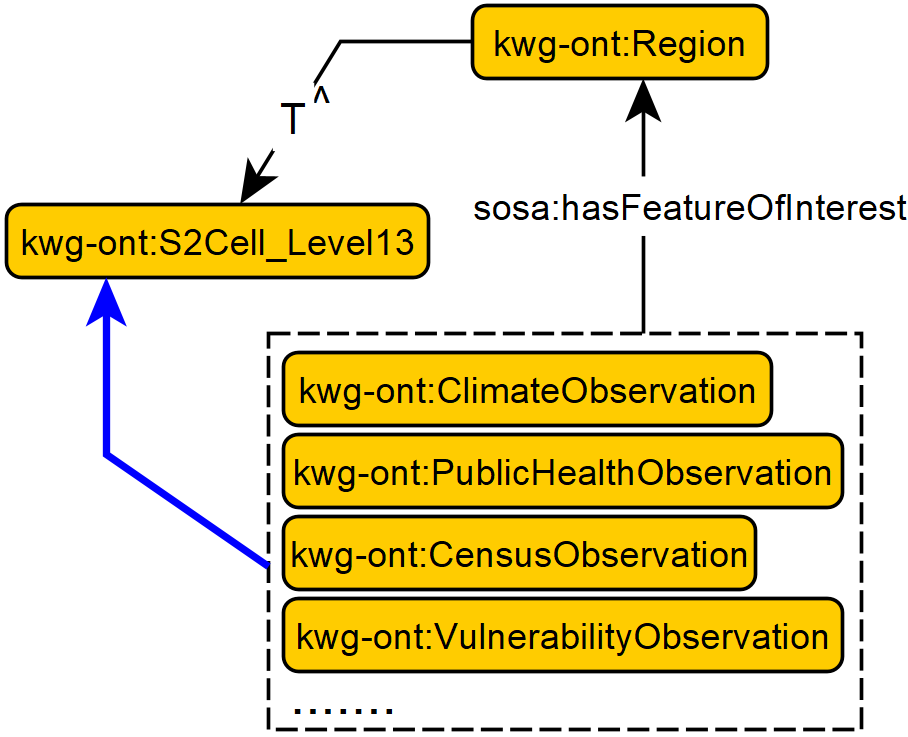}
     \caption{}
        \label{fig:indexing2}
  \end{subfigure}
  \begin{subfigure}[b]{0.45\textwidth}
    \begin{minted}[fontsize=\footnotesize, frame=lines,framesep=2mm, autogobble]{sql}
PREFIX sosa: <http://www.w3.org/ns/sosa/>
PREFIX kwg-ont: <http://stko-kwg.geog.ucsb.edu/lod/ontology/>
PREFIX time: <http://www.w3.org/2006/time#>
PREFIX xsd: <http://www.w3.org/2001/XMLSchema#>
SELECT ?fire_oc ?oc ?svi_ob WHERE {
    ?fire_oc a kwg-ont:MTBS_FireObservationCollection;
        sosa:hasFeatureOfInterest/kwg-ont:sfContains ?s;
        sosa:phenomenonTime/time:inXSDDate ?day.
    ?oc a kwg-ont:ClimateDivisionMonthlyObservationCollection;
        sosa:hasMember/sosa:hasMember ?c;
        sosa:hasFeatureOfInterest/kwg-ont:sfContains ?s.
    ?climate_ob a kwg-ont:ClimateDivisionMonthlyObservation;
        sosa:phenomenonTime/time:year ?year;
        sosa:phenomenonTime/time:month ?month.
     ?svi_ob a kwg-ont:VulnerabilityObservation;
        sosa:hasFeatureOfInterest/kwg-ont:sfContains ?s;
        sosa:hasSimpleResult ?svi. 
    ?s a kwg-ont:S2Cell_Level13.
    FILTER (?day > "2010-01-10"^^xsd:date && 
            ?day < "2010-01-31"^^xsd:date)
    FILTER (?year = "2010"^^xsd:gYear)
    FILTER (?month = "1"^^xsd:gMonth)}
    \end{minted}
    \caption{}
        \label{fig:query}
  \end{subfigure}%
  \caption{(a) Pattern for conflating observation data on S2 cells via object property paths. The blue arrow denotes the shortcut that can be computed on demand with dedicated SPARQL queries or materialized via ontology rules. (b) Sample SPARQL query to identify S2 level 13 cells linked to fire risk during January 2010.}
\end{figure}

Observation data, such as hurricane severity and impacts, population demographics, air quality indices, and public health observations, are collected through explicit or implicit sampling strategies. These datasets contain regional identifiers (e.g., ZIP code, FIPS code, NWZ code, statistical area code) but are not explicitly connected to region geometries. In KWG, these georeferenced observation data are linked to their corresponding geographic entities or regions, such as ZIP code tabulation areas, administrative region boundaries, national weather zones, and statistical areas, via the $\texttt{sosa}$:$\texttt{hasFeatureOfInterest}$ relation as shown in Figures~\ref{fig:conflation} and~\ref{fig:indexing2}.

From a query performance perspective, using S2 cells to topologically enrich geographic entities with explicit geometries is highly beneficial. This approach consequentially conflates S2 cells with various observation data via object property paths, thereby thematically contextualizing the cells. Figure~\ref{fig:conflation} shows an example of the property path from various observations to S2 cells via connected geographic regions. The result of this process is a set of mappings from observations to geographic entities to sets of S2 cells at multiple levels of the S2 grid. This thematic enrichment of S2 cells entails a level of abstraction determined by the chosen geographic region and the resolution of the cells. However, it plays a crucial role in uncovering latent patterns, for example identifying that regions with high social vulnerability scores have an increased exposure to air pollution.

\vspace{8pt}


Figure~\ref{fig:kwg-core} illustrates the ontology pattern that explicitly links observations to geographic regions using the $\texttt{sosa}$:$\texttt{hasFeatureOfInterest}$ relation. The path between observations and S2 cells is defined using OWL 2 property chains \citep{potoniec2022inductive}, specifically through the composition of $\texttt{sosa}$:$\texttt{hasFeatureOfInterest}$ and $\texttt{kwg-ont}$:$\texttt{sfOverlaps}$. In KWG, these object paths are computed on-demand via SPARQL queries.

Thematic contextualization of S2 cells provides significant advantages for use cases requiring granular spatial summarization, cross-theme data exploration, and spatial interpolation. Conflating different observational themes with S2 cells enables users to navigate geographically through thematic data spaces. This is especially valuable for locating related places within the knowledge graph. For example, an analyst from a humanitarian aid organization may search for regions based on thematic criteria, such as identifying S2 cells associated with fire forecasts, weather conditions conducive to fires, and socioeconomically vulnerable neighborhoods. This integrated data approach assists the analyst in developing effective emergency response plans. Figure~\ref{fig:triangle} illustrates the SPARQL query used to retrieve this information from KWG. Additionally, thematic conflation supports advanced geo-visualization techniques, allowing for dynamic representation of hotspots or areas of interest at different zoom levels, essential for detailed spatial analysis and informed decision-making.

\vspace{8pt}

\subsection{Multi-scale Representation of Quantitative Geospatial Data}
Quantitative datasets consist of various observation datasets typically represented as numerical values associated with specific locations. Examples include the Social Vulnerability Index (SVI) linked to counties and PM 2.5 concentrations measured at point-based air quality monitoring sites. While these datasets can be queried through conflation at specific S2 grid levels, some statistical queries may not benefit from a multi-scale spatial representation. This largely depends on the type of quantity (amounts versus measurements, as discussed in \citep{etop2024quant}). For instance, while a county-level SVI measurement can be mapped to all S2 reference cells within that county, these values cannot simply be summed or averaged to derive a value for a larger spatial region, such as a state. In contrast, the spatial overlap areas of a specific crop type from multiple S2 cells can be summed to determine the total overlap area within a larger S2 cell that contains them. The distinction between quantities that can be aggregated (mereotopological quantities) and those that cannot (arithmetic quantities) is further defined and axiomatized in \citep{etop2024quant}. While discussing the implications of discretizing both types of quantities and their aggregations over multiple resolutions is beyond the scope of this paper, understanding these distinctions is crucial for accurate geospatial analysis and data integration within the S2 Geometry framework.

When aggregating certain quantity measurements over a larger region, it is important to incorporate weighting measures to adjust for spatial biases or the uneven distribution of data points within that region. For example, an averaging query that requests the mean of data values within a specified spatial polygon can benefit from information regarding the S2 cell containing each data point, its spatial relationship to the polygon, and the size of the S2 cell. These details help determine how data points are placed within the polygon and the appropriate weights to allocate. By incorporating these weighting factors, the aggregated results may more accurately reflect the true characteristics of the entire area, rather than being skewed by the concentration or distribution of the data points. A key objective in defining such a representation is to facilitate queries at the data's coarse scale without significant loss of accuracy.

Consider the air quality observation dataset in KWG, which contains concentration measurements for pollutants such as Ozone and PM 2.5. The PM 2.5 concentration is a critical metric in air quality assessment. Typically, air quality monitoring sites are associated with counties, but their distribution within counties is neither uniform nor evenly spaced. This uneven distribution complicates the derivation of accurate county-level estimates since PM 2.5 concentrations exhibit nonlinear spatial variations. For example, urban areas require data with a spatial resolution of 200 - 400 meters to achieve acceptable modeling accuracy of pollution exposure \citep{stroh2007study}. In KWG, individual PM 2.5 measurements are mapped onto 1 x 1 km\textsuperscript{2} grid cells represented by S2 reference cells. These measurements are then hierarchically contained within lower-resolution S2 cells, which are also topologically connected to county polygon boundaries through spatial quantization. This hierarchical structuring enables users to analyze the spatial distribution of monitoring sites within counties, facilitating the calculation of appropriate weights for more accurate regional estimates. This hierarchical structuring helps users analyze the distribution of monitoring sites within counties and calculate weights for more accurate regional estimates. It accounts for the uneven density of monitoring sites, their proximity to pollution sources, and any spatial biases from uneven observation distribution. Moreover, the fine-grained overlap of monitoring sites with other data layers -- achieved through S2 enrichment and discretization -- enhances the ability to identify additional controlling factors or measures that may disproportionately influence PM 2.5 concentrations at specific locations. Thus, the combination of quantization and the hierarchical and topological relationships within the S2 grid framework significantly improves the precision and reliability of aggregating quantities over spatial scales, particularly in regions with complex spatial characteristics.

\subsection{Cross-Graph Integration} \label{cross-integration}

 \begin{figure}[h]
     \centering
     \includegraphics[width=0.6\linewidth]{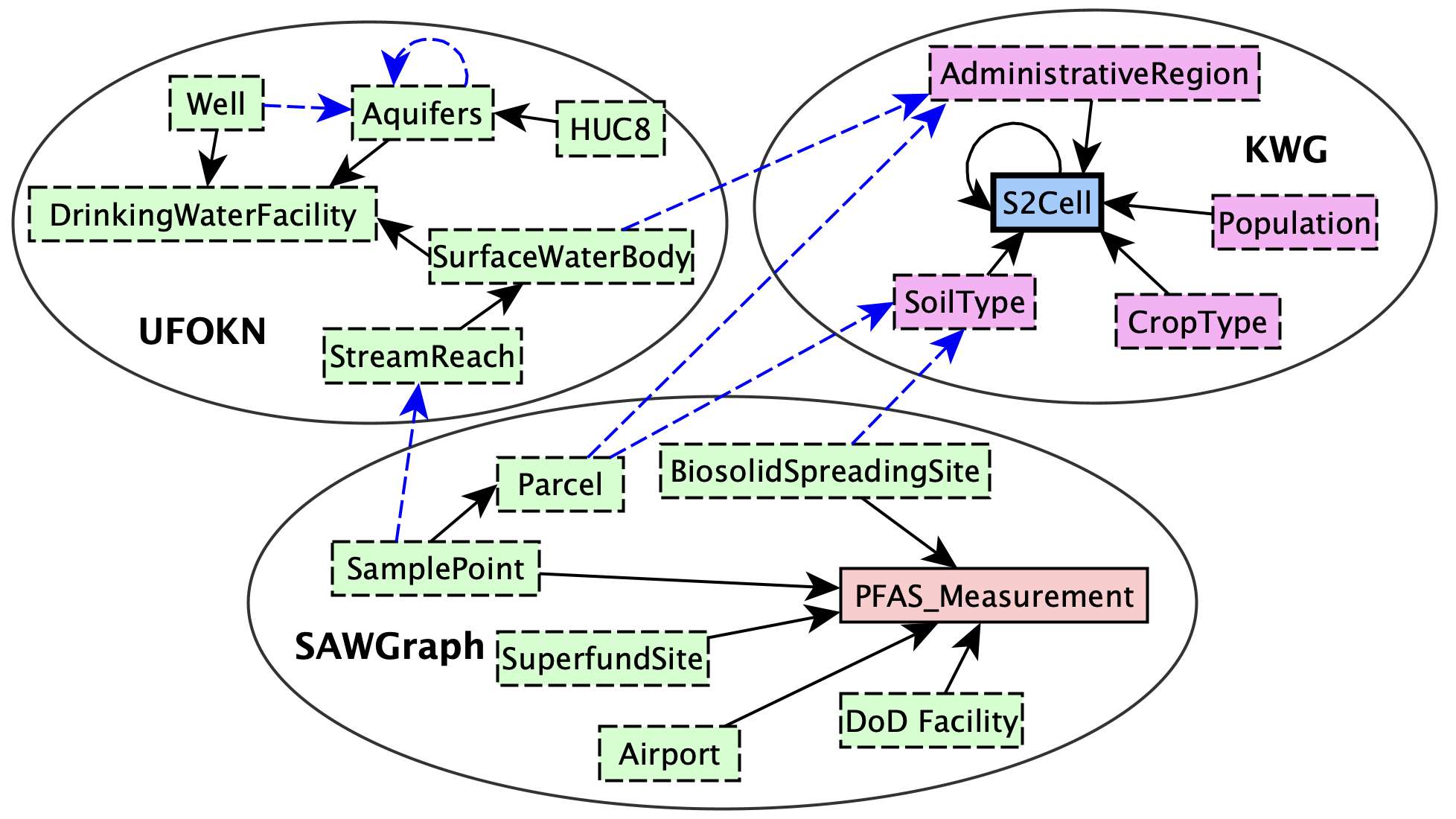}
     \caption{Overview of cross-integration between KWG, SAWGraph, and UFOKN. Pink boxes represent concepts from KWG linked to S2 cells. Solid black arrows depict prematerialized links in the graph. Green boxes, representing concepts from UFOKN and SAWGraph, are directly associated with S2 cells and conflated with other spatial and thematic concepts, as shown by the blue dashed arrows. These arrows illustrate object property paths that can be queried via SPARQL queries.}
     \label{fig:cross-links}
 \end{figure}
 
\begin{figure}[h]
    \centering
    \includegraphics[width=0.8\linewidth]{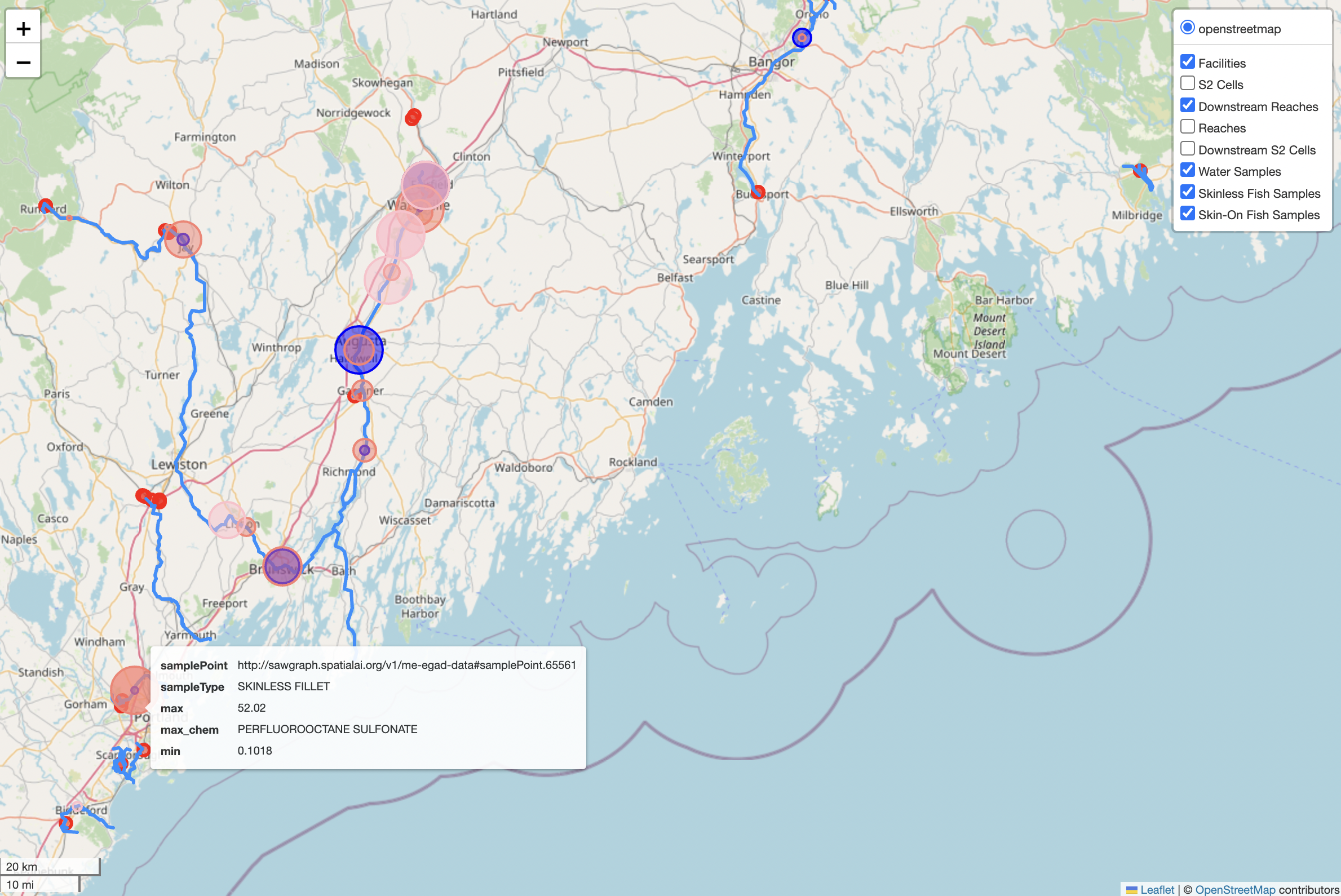}
    \caption{Screenshot of the map showing the locations of paper mills, the volume of PFAS releases, and PFAS test results in nearby or downstream water bodies and fish tissue samples. Test values are aggregated by sample location, highlighting the lowest and highest PFAS concentrations at each point. S2 cells are used to perform spatial queries, identifying locations upstream and downstream from each sample point. The Python notebook that demonstrates this query is available in this Github repository: \url{https://github.com/shirlysteph/s2-kwg/tree/main/demos/cross-graph}}
    \label{fig:saw-kwg-map}
\end{figure}

S2 serves as the \emph{spatial fabric} for cross-disciplinary graph integration and interoperability within the OKN framework, which includes SAWGraph and UFOKN, as outlined in Section~\ref{sec:background}. SAWGraph integrates datasets related to water, soil, plant and animal tissue, feed, and agricultural food products tested for per- and polyfluoroalkyl substances (PFAS), covering test results, testing locations, and potential contamination sources. This integration supports the analytical needs for PFAS monitoring and management. SAWGraph is connected with KWG and UFOKN to trace PFAS transport and accumulation pathways within the U.S. food and water systems, and to prioritize further sampling at specific locations, such as wells, agricultural lands, or crops, using federated querying with S2 cells as points of interest. To enable such functionality, each graph must implement S2 integration with relevant datasets. Figure~\ref{fig:cross-links} illustrates key spatial features from each graph that are linked to S2 cells via topological enrichment, enabling the contextualization and conflation of datasets by leveraging complementary information across graphs. Inferred links (depicted with dashed blue arrows) allow for advanced queries, such as: ``\emph{What is the relationship between PFAS releases from paper mills and PFAS test results in nearby or downstream water bodies}''. Figure~\ref{fig:saw-kwg-map} shows a map demonstrating how spatial data from the three graphs are integrated and utilized in this context. This integrated spatial framework enhances the ability to conduct cross-regional comparisons and make informed decisions regarding contamination and sampling priorities.

\vspace{8pt}

\subsection{Graph Sharding}

KWG is a continuously expanding knowledge graph, with the number of geometry nodes increasing significantly - from approximately 5.5 million in version 2.0 (KWG - Manhattan) to approximately 48 million in version 3.0 (KWG - Santa Barbara). This growth is expected to continue, necessitating careful consideration of scalability for both maintenance and performance. Managing such large volumes of data introduces challenges not only for efficient search but also for data ingestion, indexing, and updating, especially as highly dynamic environmental datasets are integrated into the graph. To mitigate the challenges posed by the increasing size, the graph can be partitioned into smaller sub-graphs or shards, which can be distributed across a cluster of servers for parallel processing. This approach leverages distributed computing to manage both the storage and query processing loads. Federated SPARQL queries can then be executed to aggregate results from the distributed shards, enabling efficient querying across the graph's full dataset while maintaining scalability and performance \citep{w3c-fed}.

\begin{figure}
    \centering
    \includegraphics[width=0.5\linewidth]{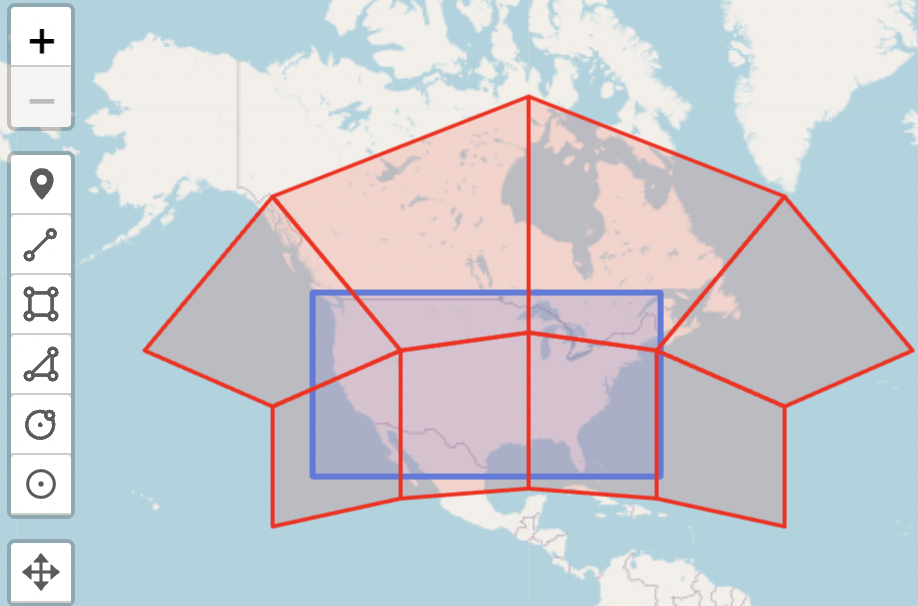}
    \caption{Eight S2 cells at level 2, drawn in red, overlap the continental U.S.}
    \label{fig:na}
\end{figure}

Several sharding approaches exist, with the appropriate method depending on factors like query patterns and data characteristics \citep{saleem2023storage}. For KWG, where the focus is on spatial insights, location-based sharding is a viable strategy, assuming spatial affinity between queries and specific regions \citep{tind-sharding}. For instance, queries might focus on population health and public health facilities in hurricane-affected areas. However, in a complex GeoKG like KWG, where data often spans fiat geographic boundaries (e.g., natural phenomena like wildfires), a purely region-based sharding approach may be insufficient. To address this, sharding should partition data in geographies with minimal cross-region interaction, reducing distributed joins during federated queries and improving efficiency while preserving spatial relationships. Geospatial data in KWG is already clustered on the S2 grid using topological indexing, conflation, and quantization, making the S2 grid a natural choice for location-based partitioning. Data from different S2 cells (at an appropriate level) can be distributed across servers, forming a cluster of shards further indexed (e.g., using Elasticsearch). For example, sharding North American data on the S2 grid at level 2 results in eight servers corresponding to the eight S2 cells overlapping the region -- see Figure~\ref{fig:na}. Shard routing algorithms, like those in Elasticsearch, can then direct federated queries to the appropriate server.


\section{Conclusion}\label{sec:conclusion}
This paper demonstrates how the implementation of S2 Geometry using GeoSPARQL's vector data model as an RDF graph establishes the DGGS as a versatile spatial data reference framework. This integration enhances the functionality of GeoKG infrastructures by eliminating the need for specialized geographic extensions typically required for raster and gridded data representation. By utilizing well-supported Simple Feature geometry types and optimized geometry serializations like WKT and GML, the integration process for heterogeneous data is streamlined. The connectivity and hierarchical relationships between S2 cells, established using GeoSPARQL's topological relations, enhance the qualitative spatial perception of the S2 RDF graph. This approach simplifies the querying of complex spatial data, reducing the need for combined vector/raster querying through cross-geometric operations and raster functions, such as raster algebra operations constructed in GeoSPARQL+ \citep{homburg2020geosparql}, by instead enabling efficient execution of spatial operations such basic SPARQL and GeoSPARQL functions.

In the KWG framework, we have successfully implemented quantization methods that treat S2 cells as vector bins for encapsulating both qualitative and quantitative geospatial information. By pre-materializing statements for this binning as RDF triples, we optimize performance for computationally expensive queries involving spatial predicates or join operations. Instead of scanning the entire graph, these queries can swiftly access relevant materialized views associated with specific S2 cells, resulting in significant improvements in query response times. Specifically, the discretization process transforms spatial data into a finer, more granular format directly linked to S2 cells, allowing for effective representation and querying of remotely sensed raster environmental observation data, which graph databases are not fully fledged to handle.

Furthermore, the geometric and semantic flexibility of the S2 RDF graph facilitates spatial graph analytics through cell-based raster-to-vector conflation. The hierarchical structure of S2 grids, along with explicit relationships between cells at different resolutions, allows for consistent multi-resolution data representation and integration of time-series geospatial data. However, we acknowledge that discretization must be approached with caution; the results' quality is highly contingent upon the original data accuracy, spatial uncertainty, and area distortion associated with each cell. By cross-linking various thematic datasets in KWG via stable, portable S2 cell identifiers, we support efficient data representation by potentially rendering some geometries redundant.

The modular data model inherent in S2 promotes flexible and efficient data management, ensuring scalability and sustainability in graph maintenance. This structure enables straightforward partitioning of spatial data, which is especially advantageous in distributed computing environments for processing large-scale geospatial datasets through parallel processing techniques. Grid cells serve as analytic tiles, facilitating client-side analytics such as machine learning model development and geovisualization.

Transforming and representing non-DGGS spatial data within a DGGS framework requires a deep understanding of the specific DGGS software library being utilized. Due to the size and complexity of geospatial datasets, particularly when dealing with intricate polygon geometries, quantization can require substantial computational power and processing time, potentially leading to oversimplifications of the original data. The prevalent use of geographic CRS such as WGS84 and EPSG:4326 in traditional geospatial workflows complicates the accurate handling of spherical geometries like those used by the S2 library. Accurate conversion between common geographic CRS and spherical representations is critical, particularly when dealing with large spatial extents, as transitioning from vector or raster formats to DGGS cells can lead to data loss or inaccuracies. Moreover, the variable resolution of DGGS cells can introduce issues such as the Modifiable Areal Unit Problem, as not all S2 cells are equal-area at each level. When discretizing information onto S2 cells, it is essential to carefully quantify and express spatial uncertainty, especially when working across multiple scales. While encoding spatial uncertainty within DGGS cells enhances the overall accuracy of geospatial analytics, it also introduces complexities for decision-makers who are accustomed to working with traditional raster and vector formats. Moreover, this shift raises important questions about how DGGS impacts multiscale analysis and its suitability for decision-making workflows that rely on high-fidelity geographic data

This paper outlines a comprehensive methodological framework for leveraging S2 Geometry as an integration framework, enabling the semantic reconciliation and fusion of diverse vector and raster datasets across multiple spatial scales and CRS. By addressing the limitations of traditional graph databases in handling complex geospatial data, our approach introduces quantization techniques within KWG that are tailored to specific analytical use cases. These methods not only data accessibility but also support a broader vision of advanced spatial analytics within the context of GeoKGs. While we acknowledge current challenges -- such as spatial uncertainty and computational scalability -- our goal is to foster collaboration within the geographic information science research community by presenting critical open questions. This dialogue is essential for advancing more accurate, integrated geospatial analyses and fostering innovation within the evolving GeoKG landscape.

\bibliographystyle{unsrtnat}
\bibliography{template}  






\end{document}